%% file: main.tex
\pdfoutput=1 
\documentclass{article}

\newif\ifpreprint
\preprinttrue

\usepackage{iclr2027_conference,times}
\ifpreprint
  \iclrfinalcopy 
\fi

\usepackage[utf8]{inputenc}
\usepackage[T1]{fontenc}
\usepackage{hyperref}
\usepackage{url}
\usepackage{booktabs}
\usepackage{placeins}
\usepackage{enumitem}
\usepackage{amsfonts}
\usepackage{nicefrac}
\usepackage{microtype}
\usepackage{xcolor}

\usepackage{graphicx}
\usepackage{wrapfig}
\usepackage{subcaption}
\usepackage{fvextra} 
\usepackage{amsmath}
\usepackage{amssymb}
\usepackage{mathtools}
\usepackage{amsthm}
\usepackage[capitalize,noabbrev]{cleveref}
\usepackage{makecell}
\usepackage{tcolorbox}
\usepackage{natbib}
\input{def}

\usepackage[disable,textsize=tiny]{todonotes}

\title{Specification-aware Predictive Uncertainty Decomposition}

\author{%
  Roman Vashurin$^{1}$,
  Mikhail Seleznyov$^{2,5}$,
  Assel Yermekova$^{3}$,
  Fedor Chernogorskii$^{1}$,\\
  \textbf{Danila Zykov$^{4}$,
  Elena Tutubalina$^{4}$,
  Alexander Panchenko$^{2,5}$,
  Preslav Nakov$^{1}$,}\\
  \textbf{Nikita Kotelevskii$^{1}$,
  Maxim Panov$^{1}$}\\[0.5ex]
  $^{1}$MBZUAI\quad
  $^{2}$AXXX\quad
  $^{3}$Huawei Technologies Ltd.\quad
  $^{4}$HSE\quad
  $^{5}$S-NLP\\
  \texttt{Roman.Vashurin@mbzuai.ac.ae}
}

\begin{document}

\maketitle
\ifpreprint
  \fancyhead{}
  \renewcommand{\headrulewidth}{0pt}
\fi

\begin{abstract}
Classical decompositions of predictive uncertainty, such as the information-theoretic one, are defined for a fixed representation \(x\) of the input object. They do not consider that \(x\) may be incomplete, nor do they measure how much the missing information can affect the prediction.
In Large Language Models, users can clarify an underspecified input without any retraining. The clarified input can describe the user's unobserved intent \(z^*\) more accurately and change the predictive uncertainty about the outcome.
In this paper, we formalize this setting probabilistically and derive a three-term decomposition of prediction risk into \emph{specification}, residual aleatoric, and epistemic uncertainties.
We show that at the original information level, when only \(x\) is available, the aleatoric uncertainty, represented by a Bayes risk, has two components. The first, which we call specification uncertainty, is the expected reduction in this Bayes risk from revealing the intent. The second is the residual aleatoric uncertainty, i.e., the Bayes risk that remains once the intent is known, averaged over its possible values \(z\) given \(x\).
The distinction is important, as specification uncertainty informs the decision to request clarification before making a prediction.
We build approximations of these components and connect them to existing uncertainty measures. Our experiments across different domains and modalities show distinct roles for the three components.
Specification uncertainty is the most informative for detecting underspecified inputs, while residual aleatoric uncertainty is the most informative for detecting in-distribution prediction errors.
\end{abstract}

\input{sections/introduction}
\input{sections/related_work}
\input{sections/method}

\input{sections/results}
\input{sections/conclusion}


\bibliography{main}
\bibliographystyle{iclr2027_conference}

\newpage
\appendix   

\input{sections/appendix}

\end{document}

%% file: def.tex
\newcommand{\RR}{\mathbb{R}}
\newcommand{\EE}{\mathbb{E}}
\newcommand{\HH}{\mathbb{H}}

\newcommand{\X}{\mathcal{X}}
\newcommand{\Y}{\mathcal{Y}}
\newcommand{\Z}{\mathcal{Z}}
\newcommand{\Pcal}{\mathcal{P}}

\newcommand{\scalarprod}[2]{\langle #1 \, , \, #2 \rangle}

\theoremstyle{plain}

\theoremstyle{definition}

\theoremstyle{remark}


%% file: sections/introduction.tex

\section{Introduction}
\label{sec:intro}

\begin{figure}[t]
    \centering
    \includegraphics[width=0.75\linewidth]{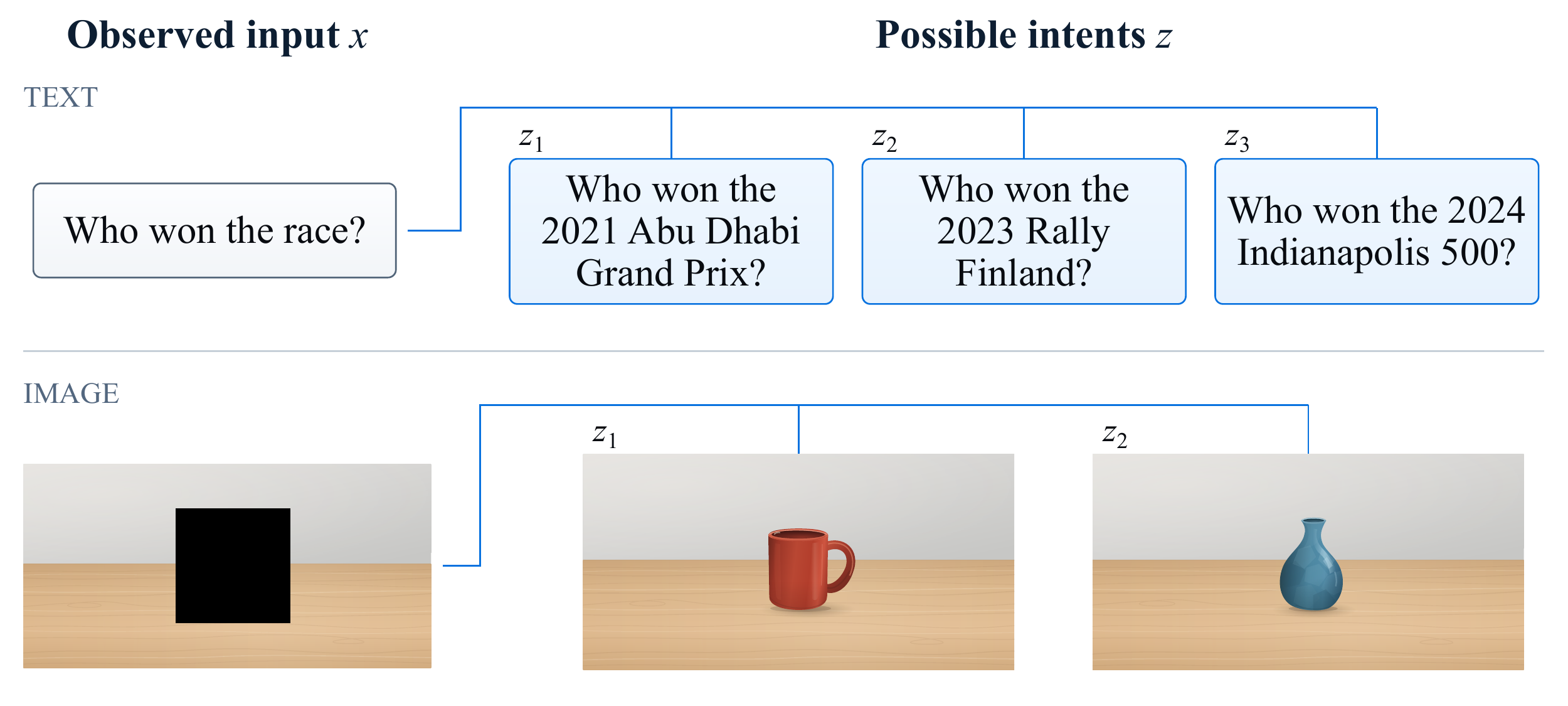}
    \caption{Examples of observed inputs \(x\) (the ones provided by users) and possible values of the unobserved intent \(z\). Different intended questions (top) or complete images (bottom) can be consistent with the same observed input. The actual value \(z^*\) remains unobserved.}
    \label{fig:x_z_completion_examples}
\end{figure}

As artificial intelligence (AI) systems, especially generative ones, are widely adopted~\citep{bick2026rapid}, there is a growing demand for methods and practices to control their reliability.
One such direction is predictive Uncertainty Quantification (UQ; \citealp{hullermeier2021aleatoric}), which supplements the original model's output with a value (typically a scalar) describing the model's certainty in its prediction.
Based on this value, a practitioner can make a practical decision, such as to abstain from the prediction, request additional information, or proceed with the prediction if the uncertainty is acceptable~\citep{geifman2017selective,mozannar2020consistent,naszadi2023aligning}.
For Large Language Models (LLMs)~\citep{brown2020language}, requesting additional information is particularly natural, since a user can clarify the input without retraining the model.
For example, given the prompt ``Who won the race?'', a model may either answer right away or first ask which race is meant (see Figure~\ref{fig:x_z_completion_examples}).
Therefore, uncertainty in prediction may come from different \emph{sources}, and different sources may call for different actions~\citep{sale2026aleatoric}.

Two \emph{types} of uncertainty are widely distinguished~\citep{hullermeier2021aleatoric}. \emph{Aleatoric} uncertainty is irreducible with respect to the considered informational constraints, while \emph{epistemic} uncertainty is reducible and characterizes the lack of knowledge about the underlying ground-truth data-generation process.
Different formalizations of these types lead to concrete UQ measures~\citep{gal2016dropout,kendall2017uncertainties,lakshminarayanan2017simple,kotelevskii2025risk,kotelevskii2022nonparametric,sale2023volume,sale2024second,hofman2024quantifying,schweighofer2024information}.

For LLMs, many UQ methods have been proposed~\citep{lin2024generating,farquhar2024detecting,tian-etal-2023-just,vashurin2025cocoa}, including decompositions into aleatoric and epistemic uncertainty~\citep{malinin2021uncertainty,ma2025estimating}.
However, these decompositions are defined for a given input representation and do not ask whether clarifying the input could improve the prediction.

Recent work considered uncertainty arising from \emph{underspecified} inputs~\citep{input_clarification,walha2026finegraineduncertaintydecompositioninlargelanguagemodelsaspectralapproach}.
These approaches generate clarifications of the input and treat the variability of predictions across them, quantified by Bregman information, as aleatoric uncertainty, and the expected uncertainty remaining after clarification as epistemic uncertainty.
However, the outcome can be random even for a fully clarified input, so the remaining uncertainty is not necessarily epistemic.
Moreover, both approaches use a single fixed model and thus ignore uncertainty about its parameters, which the information-theoretic decomposition treats as epistemic.

We present an alternative view on this distinction.
We assume that each observed input \(x\) (possibly only partially) describes the true intent~\(z^*\), which itself is never observed, and ask how much the prediction would improve by revealing this intent.
Measuring uncertainty by the expected loss of a prediction (also called risk), we obtain a three-term decomposition of this risk into \emph{Specification}, \emph{Epistemic}, and \emph{residual Aleatoric} uncertainty (SEA).
Specification uncertainty is the expected reduction in Bayes risk (the risk of the data-generation process at \(x\)) from revealing the intent, that is, the expected gain from clarification. When it is large, it is worth requesting clarification before providing an answer.
At the original information level (i.e., when only \(x\) is available), specification uncertainty is \emph{only a part} of the aleatoric uncertainty.

We summarize our contributions below.

\begin{itemize}
    \item We derive a new UQ decomposition, relevant in particular for LLMs, that splits the predictive uncertainty into three terms by separating specification uncertainty from both the residual aleatoric uncertainty and the epistemic one. When the input fully specifies the intent, the specification term vanishes, and the decomposition reduces to the risk-based one~\citep{kotelevskii2025risk}, generalizing the information-theoretic uncertainty decomposition; see Section~\ref{sec:decomposition-theory}.

    \item For a specific class of loss functions, proper scoring rules, we derive closed-form uncertainty measures and, following the risk-based approach, their Bayesian approximations. These approximations generalize the clarification-based decompositions previously derived for a single fixed model~\citep{input_clarification,walha2026finegraineduncertaintydecompositioninlargelanguagemodelsaspectralapproach}; see Section~\ref{sec:bayesian-approximations}.
    
    \item In our experiments, we demonstrate that each of the derived components is well-suited for a particular type of decision in downstream problems. Specification uncertainty is the most informative of the three components for detecting underspecified inputs, while residual aleatoric uncertainty is the most informative component for detecting prediction errors for in-distribution data; see Section~\ref{sec:results} and Appendix~\ref{app:image-classification}.
\end{itemize}

%% file: sections/related_work.tex
\section{Related Work}

\textbf{Predictive uncertainty decomposition.}
There are several approaches to formalize the decomposition into aleatoric and epistemic uncertainty and estimate the corresponding components.
The information-theoretic approach considers a Bayesian (or, more generally, second-order) model. It associates the conditional mutual information between predictions and model parameters with epistemic uncertainty, and the expected entropy of the prediction given the parameters with aleatoric uncertainty~\citep{houlsby2011bayesianactivelearning}. Estimating the epistemic term, therefore, requires a second-order learner, such as a Bayesian model or an ensemble~\citep{sale2023volume,sale2024second}.
The risk-based formulation generalizes this approach. It decomposes the pointwise prediction risk into the Bayes risk (aleatoric uncertainty) and the excess risk (epistemic uncertainty). These components can be approximated with Bayesian models~\citep{kotelevskii2025risk,schweighofer2024information}, which recover the decomposition of \citet{houlsby2011bayesianactivelearning} as a special case. Another strategy for approximating these terms uses specialized deterministic auxiliary models~\citep{kotelevskii2022nonparametric,lahlou2021deup}.
Other methods estimate uncertainty with a single deterministic network by modifying its architecture or fitting a density model on its features~\citep{van2020uncertainty,liu2020simple,mukhoti2021deep}.
For autoregressive generation, the information-theoretic decomposition has been extended to the token and sequence levels~\citep{malinin2021uncertainty}, still requiring a Bayesian model or an ensemble for estimation.
Another approach, LogTokU~\citep{ma2025estimating}, estimates aleatoric and epistemic uncertainties from logits without repeated sampling by interpreting them as evidence for a Dirichlet distribution.

These decompositions quantify uncertainty at a fixed level of input information, when the representation of an input object (or its prompt) \(x\) is given. They do not consider how the risk and the predictive uncertainty would change if \(x\) were clarified.
Interactive LLMs, however, allow users to clarify their inputs without retraining the model, which makes this question natural.

\textbf{Specification uncertainty.}
\citet{sale2026aleatoric} argue that the aleatoric/epistemic distinction depends on the information available in a given setting, for example, on whether an ambiguous prompt can be clarified.
\citet{input_clarification} propose input-clarification ensembling, which decomposes predictive uncertainty using generated clarifications of the prompt. In their decomposition, the mutual information between the prediction and the clarification is treated as aleatoric uncertainty caused by input ambiguity, while the expected entropy of the response given a clarification is treated as epistemic uncertainty.
\citet{walha2026finegraineduncertaintydecompositioninlargelanguagemodelsaspectralapproach} keep this interpretation and generalize it with functional Bregman information, instantiated with kernel-based von Neumann entropy.
Section~\ref{sec:method} shows that their decomposition is a special case of ours for a single fixed model and that its terms estimate specification and residual aleatoric uncertainty.
In contrast, we split the Bayes risk at the level of \(x\) into two parts. Specification uncertainty is the expected reduction in Bayes risk from revealing the intent, and residual aleatoric uncertainty is the expected Bayes risk after the intent is revealed. The excess risk of the predictor is still treated as epistemic uncertainty.

%% file: sections/method.tex

\section{Risk Decomposition with Unobserved Intents}
\label{sec:method}

In this work, we consider a setup in which the observed representation \(x \in \X\) of an object may be incomplete and can be clarified by additional information.
Our goal is to quantify how much such a clarification is expected to improve the prediction, and to separate this gain from other sources of predictive uncertainty.

In what follows, by \emph{intent}, we mean a fully specified input that \(x\) describes only partially. We denote the user's true but unobserved intent by \(z^* \in \Z\).
For example, a user may have a specific question \(z^*\) in mind but write a less specific prompt \(x\). Similarly, \(x\) could be a masked image or an audio recording with a missing interval, and \(z^*\) the original image or the full recording.
Several intents can be consistent with the same \(x\) (see Figure~\ref{fig:x_z_completion_examples}), and the conditional distribution \(p(z\mid x)\) describes which values \(z^*\) is likely to take.
In these examples, \(\Z = \X\), since a clarified prompt is again a prompt, and a completed image or recording has the same format as the masked one. Our framework, however, does not require the two spaces to coincide.

We assume that clarification does not change the output space \(\Y\), as otherwise the prediction risks before and after clarification could not be compared. For a classification task, this means that the set of output classes must be the same for \(x\) and \(z\).
Finally, we assume that once \(z\) is known, \(x\) provides no additional information about \(Y\). Formally, \(Y \perp X \mid Z\), i.e., \( p(y \mid x, z) = p(y \mid z) \).

Under these assumptions, we consider the following \emph{ground-truth} process
\begin{equation}
    p(y, z \mid x) = p(y \mid z) \, p(z \mid x).
    \label{eq:process_definition}
\end{equation}
Note that, for uncertainty quantification, we consider a distribution already conditioned on a specific \(x\), as we are typically interested in evaluation at a particular point.
In what follows, we write \(p_z\coloneqq p(y\mid z)\) and \(p\coloneqq p(y\mid x) = \EE_{Z\sim p(z\mid x)} \, p_Z\).

\subsection{Three-component decomposition of the expected risk}
\label{sec:decomposition-theory}

We follow the formalization of uncertainty as the expected loss (or risk) of a prediction~\citep{kotelevskii2025risk}. Namely, if we expect that, for whatever reason, the model has a large loss at a given input \(x\), we say that its prediction at \(x\) is uncertain.
Below, we consider a specific class of loss functions, proper scoring rules~\citep{gneiting2007strictly}. For such losses, the expected loss is minimized by the true distribution of the outcome, which lets us separate the risk of the best possible prediction from the excess risk of the model.

Let \( \ell(q, y) \) denote a proper scoring rule~\citep{gneiting2007strictly} that evaluates a predictor \(q \coloneq q(y \mid x) \) at \(Y\).
We define the expected pointwise risk of \(q\) as
\begin{equation*}
    R(p,q) \coloneqq \EE_{Y\sim p}[\ell(q,Y)].
\end{equation*}

\textbf{Decomposing to Bayes risk and excess risk.}
We can decompose this expected risk into the Bayes risk \(R_b(p)\coloneqq R(p,p)\) of the process \(p\) and the excess risk \(R_{\mathrm{exc}}(p~\|~q)\coloneqq R(p,q)-R(p,p)\) of the model \(q\) with respect to \(p\).
Following~\citet{kotelevskii2025risk}, for a proper scoring rule instantiated with a convex function \( G \colon \Pcal(\Y) \to \RR \), this gives
\begin{equation}
    R(p,q) = R_b(p)+R_{\mathrm{exc}}(p~\|~q) = -G(p)+D_G(p~\|~q),
\label{eq:proper-scoring-rules-decomposition-risk}
\end{equation}
where \(D_G(p~\|~q) = G(p) - G(q) - \scalarprod{G'(q)}{p-q}\) is the Bregman divergence instantiated by \(G\).
Here, \(-G(p)\) is the Bayes risk. It depends only on the data generation process \(p\), so changing the model does not affect it. This term corresponds to aleatoric uncertainty, which is \emph{irreducible}. The excess risk \(D_G(p~\|~q)\) depends on the model, its training data, and the training procedure. It can therefore be reduced by improving the model and corresponds to epistemic uncertainty, which is \emph{reducible}.
Reducibility here means that the predictor can be improved while the available input information is kept fixed.

\textbf{Splitting the Bayes risk over intents.}
Interestingly, the Bayes risk \(-G(p)\) can be split further by using the ground-truth process~\eqref{eq:process_definition}. Since the true intent \(z^*\) is unobserved, we treat it as a random variable \(Z \sim p(z \mid x)\).
Then, using \(p_z\) and \(p\) defined above, we can expand equation~\eqref{eq:proper-scoring-rules-decomposition-risk} as follows (see Appendix~\ref{app:bayes-risk-split} for the derivation):
\begin{equation}
    R(p,q)
    = \overbrace{\underbrace{\EE_{Z \sim p(z \mid x)}[-G(p_Z)]}_{\text{residual aleatoric}}
    + \underbrace{\EE_{Z \sim p(z \mid x)}[D_G(p_Z~\|~p)]}_{\text{specification}}}^{-G(p)}
    + \underbrace{D_G(p~\|~q)}_{\text{epistemic}}.
    \label{eq:gt_risk_decomposition}
\end{equation}

\textbf{Interpretation of the three components.}
As in prior work~\citep{kotelevskii2025risk,kotelevskii2022nonparametric,lahlou2021deup}, the aleatoric uncertainty in equation~\eqref{eq:proper-scoring-rules-decomposition-risk} is defined relative to the input \(x\).
\citet{sale2026aleatoric} argue that the split into reducible and irreducible uncertainty depends, among other things, on the features available as input.
In particular, uncertainty that is irreducible given \(x\) can still be reduced by providing more input information, such as the intent.
Therefore, both \( p(y \mid x) \) and \( p(y \mid z) \) determine ground-truth aleatoric uncertainty, \emph{but relative to different fixed information levels.}

For the joint process \(p(y, z \mid x)\) that we assume here, \(p :=  p(y \mid x) \) and \(p_z :=  p(y \mid z) \) describe the predictive distribution before and after the intent is revealed.
Namely, the best prediction based only on \(x\) is \(p\), and its Bayes risk is \(-G(p)\).
If the intent is revealed and turns out to be \(z\), the best prediction is \(p_z\), with Bayes risk \(-G(p_z)\).
Thus, before the intent is revealed, the expected Bayes risk of a predictor that will observe it is \(-\EE_{Z\sim p(z\mid x)}[G(p_Z)]\). We call this term \emph{residual aleatoric uncertainty}.

The difference between these two Bayes risks is the Bregman information term in equation~\eqref{eq:gt_risk_decomposition}:
\begin{equation*}
    \underbrace{-G(p)}_{\text{before clarification}}
    - \underbrace{\big(-\EE_{Z\sim p(\cdot\mid x)}[G(p_Z)]\big)}_{\text{after clarification}}
    = \EE_{Z\sim p(\cdot\mid x)}\big[D_G(p_Z\|p)\big].
\end{equation*}
It shows how much the Bayes risk is expected to decrease once the intent is revealed, i.e., how much we expect to gain from clarification. We call this term \emph{specification uncertainty}.
Like the Bayes risk, specification uncertainty depends only on the ground-truth process.
If \(x\) fully specifies the intent (i.e., \(p(z\mid x)\) is degenerate), specification uncertainty vanishes, and equation~\eqref{eq:gt_risk_decomposition} reduces to equation~\eqref{eq:proper-scoring-rules-decomposition-risk}.

Each term in equation~\eqref{eq:gt_risk_decomposition} has a clear practical meaning, as it corresponds to a specific action.
In expectation, clarification reduces the Bayes risk through the specification term, and improving the predictor reduces the epistemic term.
Neither action reduces the residual aleatoric term, so the only way to avoid this part of the risk is to abstain from prediction.

\subsection{Bayesian approximation of the components}
\label{sec:bayesian-approximations}
In practice, the ground-truth process \(p(y, z \mid x)\) in equation~\eqref{eq:process_definition} is unknown, and we need to approximate it.
For this, we consider a model \(q(y \mid x, \theta)\), parametrized by \(\theta \in \Theta\).
To keep generality and show the connection with known decompositions in the literature, we assume this model is Bayesian:
\begin{equation*}
    q(y,z,\theta \mid x, D) = q(y\mid z, \theta) \, \pi(\theta, z \mid x, D),
\end{equation*}
where \( D = \{(x_i, y_i)\}_{i=1}^N \) is the training dataset.
The model has two latent variables, the parameters \(\theta\) and the intent \(z\), and the joint distribution \(\pi(\theta, z \mid x, D)\) is their posterior.

The components of equation~\eqref{eq:gt_risk_decomposition} depend on the unknown distributions \(p(z\mid x)\) and \(p_z\). To obtain computable and practical uncertainty measures, we replace them with approximations. We describe how we obtain them at the end of this section.

These approximations can be incorporated in several ways \citep{kotelevskii2025risk}, and we discuss them in Appendix~\ref{app:other_approximations}.
Here in the main part, we use the one that recovers the standard information-theoretic decomposition, as we show below.
It treats each posterior sample of the parameters as a candidate for the ground-truth process and uses the posterior predictive distribution as the predictor.
Let \(q_{z,\theta}\coloneqq q(y\mid z,\theta)\), and define
\begin{equation*}
    q \coloneqq \EE_{\pi}[q_{Z,\Theta}] = q(y\mid x,D), \qquad
    q_\theta \coloneqq \EE_{\pi}[q_{Z,\Theta}\mid\Theta=\theta],
\end{equation*}
where the expectations are computed under \(\pi(\theta,z \mid x,D)\).
The distribution \(q\) is the posterior predictive distribution of this model and specifies the predictor introduced above.
Averaging the risk over the candidates \(q_\theta\), we estimate \(R(p,q)\) by \(\EE_\Theta[R(q_\Theta,q)]=-G(q)\).
Decomposing the Bayes risk of each \(q_\theta\), we obtain estimates of the three components of equation~\eqref{eq:gt_risk_decomposition}:
\begin{equation}
    -G(q)
    = \underbrace{-\EE_{Z,\Theta}[G(q_{Z,\Theta})]}_{\text{residual aleatoric}}
    + \underbrace{\EE_{Z,\Theta}[D_G(q_{Z,\Theta}~\|~q_\Theta)]}_{\text{specification}}
     + \underbrace{\EE_\Theta[D_G(q_\Theta~\|~q)]}_{\text{epistemic}}.
\label{eq:our_decomposition_general}
\end{equation}

For \( G(q) = \sum_i q_i \log q_i \), equation~\eqref{eq:our_decomposition_general} becomes
\begin{equation}
\begin{aligned}
    \HH(q)
    &=
    \overbrace{
        \underbrace{\EE_{Z,\Theta}[\HH(q_{Z,\Theta})]}_{\mathrm{SEA}_{\mathrm{res}}}
        + \underbrace{\EE_{Z,\Theta}[KL(q_{Z,\Theta}~\|~q_\Theta)]}_{\mathrm{SEA}_{\mathrm{spec}}}
    }^{\text{aleatoric uncertainty estimate}}
    + \underbrace{\EE_\Theta[KL(q_\Theta~\|~q)]}_{\mathrm{SEA}_{\mathrm{epi}}} \\
    &= \EE_\Theta[\HH(q_\Theta)]
       + I(Y;\Theta\mid X=x,D),
\end{aligned}
\label{eq:specific_bald_decomposition}
\end{equation}
where the second equality holds because \(q_\theta\) averages \(q_{z,\theta}\) over intents, so the first two terms sum to \(\EE_\Theta[\HH(q_\Theta)]\).
The second line is therefore the standard information-theoretic predictive uncertainty decomposition, applied to the model with the intent marginalized out. Its epistemic term is the conditional mutual information, which is used as the BALD acquisition score~\citep{houlsby2011bayesianactivelearning}.
Below, we denote the three terms by \(\mathrm{SEA}_{\mathrm{res}}\), \(\mathrm{SEA}_{\mathrm{spec}}\), and \(\mathrm{SEA}_{\mathrm{epi}}\), and their sum \(\HH(q)\) by \(\mathrm{SEA}_{\mathrm{total}}\).
The standard decomposition merges \(\mathrm{SEA}_{\mathrm{res}}\) and \(\mathrm{SEA}_{\mathrm{spec}}\) into one aleatoric term, so it cannot show whether clarifying \(x\) would reduce the risk.
We discuss limitations of our approach in Appendix~\ref{app:limitations}.

\textbf{Connection to clarification-based decompositions.}
Our work is closely related to \citep{input_clarification} and \citep{walha2026finegraineduncertaintydecompositioninlargelanguagemodelsaspectralapproach}, which both consider a single fixed model and generate clarifications of its input.
If we set \( \theta = \text{const}\) and write \(q_z \coloneqq q_{z,\theta}\), then equation~\eqref{eq:our_decomposition_general} simplifies to
\begin{equation}
    -G(q) = \underbrace{-\EE_{Z}[G(q_{Z})]}_{(ii)} + \underbrace{\EE_{Z}[D_G(q_{Z}~\|~q)]}_{(i)}.
    \label{eq:single_model_general_decomposition}
\end{equation}
A specific choice of \( G \) gives equation 9 of \citep{walha2026finegraineduncertaintydecompositioninlargelanguagemodelsaspectralapproach}.
For \( G(q) = \sum_i q_i \log q_i \), we get \(\HH(q) = \EE_{Z}[KL(q_{Z}~\|~q)] + \EE_{Z}[\HH(q_{Z})]\), which is exactly equation 3 of \citep{input_clarification}.

Both works interpret term (i) as aleatoric uncertainty and term (ii) as epistemic uncertainty.
To see what these terms measure, note that equation~\eqref{eq:single_model_general_decomposition} holds for any predictor, including the ideal one with \(\pi(z\mid x) = p(z\mid x)\) and \(q_z = p_z\) for all \(z\).
Then \(q = p\), the excess risk \(D_G(p~\|~q)\) vanishes, and equation~\eqref{eq:single_model_general_decomposition} becomes
\begin{equation*}
    -G(p) = -\EE_{Z}[G(p_{Z})] + \EE_{Z}[D_G(p_{Z}~\|~p)],
\end{equation*}
which is exactly the split of the Bayes risk in equation~\eqref{eq:gt_risk_decomposition}.
The crucial consequence is that term (ii) is nonzero whenever the outcome is random given the intent, even \emph{if we have perfect knowledge} of \(p(y\mid z)\) and \(p(z\mid x)\).
Hence, this term becomes irreducible, and treating it as epistemic uncertainty would suggest improving a predictor that is already perfect, while the appropriate action is to abstain.
Therefore, the interpretation of \citet{input_clarification} and \citet{walha2026finegraineduncertaintydecompositioninlargelanguagemodelsaspectralapproach} is justified only when the outcome is a deterministic function of the intent.
In this case, for the ideal predictor, the residual aleatoric term (ii) vanishes, and term (i) recovers the whole aleatoric uncertainty.

One may argue that for a single fixed model, the Bayesian estimate of the excess risk is zero, so equation~\eqref{eq:single_model_general_decomposition} already accounts for all uncertainty that such a model can express, and term (ii) is the only term that can represent its epistemic uncertainty.
However, as shown above, term (ii) estimates the residual aleatoric uncertainty.
The zero estimate of the excess risk only means that a point-mass posterior cannot express epistemic uncertainty, not that the model has none.
Such uncertainty can instead be estimated by non-Bayesian methods~\citep{kotelevskii2022nonparametric,lahlou2021deup}.

\input{tables/text_qa_main_results}
\textbf{Estimation in practice.}
The pairs \((x,y)\) alone determine \(p(y\mid x)\), but not how its Bayes risk \(-G(p)\) splits into specification and residual aleatoric uncertainty.
For example, the same \(p(y\mid x)\) can arise when \(x\) fully specifies the intent and specification uncertainty is zero. It can also arise when \(y\) is a deterministic function of \(z\), and then all the randomness of \(y\) given \(x\) comes from the unknown intent.
Therefore, the two terms can be estimated separately only with additional information about the intent or with assumptions on the latent process.
In our experiments, this information comes from large pretrained generative models.
We use them to generate possible intents, which serve as hypotheses about \(z^*\). Since this is done independently of \(\theta\), we have \(\pi(\theta, z \mid x, D) = \pi(\theta \mid D) \, \pi(z \mid x)\).
Since these intents are only hypotheses, the quality of the estimates depends on how well \(\pi(z\mid x)\) approximates the true \(p(z\mid x)\).
Similarly, the ensembles that we use as samples of \(\theta\) only approximate the posterior \(\pi(\theta\mid D)\), especially when they consist of different pretrained models.

%% file: tables/text_qa_main_results.tex
\begin{table}[t!]
\centering
\small
\setlength{\tabcolsep}{3pt}
\renewcommand{\arraystretch}{1.1}
\caption{Text QA underspecification-detection AUROC using semantic entropy. CoQAR-U contains 1,000 matched damaged/undamaged pairs, with damaged inputs positive. AmbigQA contains 1,971 unique questions: 1,141 ambiguous positives and 830 unambiguous negatives.}
\label{tab:text_qa_main_results}
\begin{tabular*}{\linewidth}{@{\extracolsep{\fill}}lcccccc@{}}
\toprule
 & \multicolumn{2}{c}{Gemma E2B} & \multicolumn{2}{c}{Gemma E4B} & \multicolumn{2}{c}{Heterogeneous} \\
\cmidrule(lr){2-3}\cmidrule(lr){4-5}\cmidrule(l){6-7}
Method & CoQAR-U & AmbigQA & CoQAR-U & AmbigQA & CoQAR-U & AmbigQA \\
\midrule
$\mathrm{SEA}_{\mathrm{res}}$   & 0.5373 & 0.5102 & 0.5760 & 0.5071 & 0.5404 & 0.5350 \\
$\mathrm{SEA}_{\mathrm{epi}}$   & 0.5213 & 0.5442 & 0.5303 & 0.5461 & 0.5232 & 0.4396 \\
$\mathrm{SEA}_{\mathrm{spec}}$  & \textbf{0.7807} & \textbf{0.6575} & \textbf{0.7709} & \textbf{0.6678} & \textbf{0.7873} & \textbf{0.6017} \\
$\mathrm{SEA}_{\mathrm{total}}$ & 0.7365 & 0.5169 & 0.7614 & 0.5164 & 0.6593 & 0.5250 \\
\midrule
\multicolumn{7}{@{}l}{BALD (without clarifications)} \\
$\mathrm{BALD}_{\mathrm{alea}}$  & \textbf{0.7356} & 0.5011 & 0.7607 & 0.4858 & 0.7706 & \textbf{0.5073} \\
$\mathrm{BALD}_{\mathrm{epi}}$   & 0.6002 & \textbf{0.5442} & 0.6081 & \textbf{0.5348} & 0.7109 & 0.4535 \\
$\mathrm{BALD}_{\mathrm{total}}$ & 0.7340 & 0.5015 & \textbf{0.7608} & 0.4865 & \textbf{0.7804} & 0.5003 \\
\midrule
\multicolumn{7}{@{}l}{BALD (clarification posterior)} \\
$\mathrm{BALD}_{\mathrm{alea}}$  & \textbf{0.7413} & 0.5169 & \textbf{0.7627} & 0.5162 & \textbf{0.6721} & \textbf{0.5383} \\
$\mathrm{BALD}_{\mathrm{epi}}$   & 0.5213 & \textbf{0.5442} & 0.5303 & \textbf{0.5461} & 0.5232 & 0.4396 \\
$\mathrm{BALD}_{\mathrm{total}}$ & 0.7365 & 0.5169 & 0.7614 & 0.5164 & 0.6593 & 0.5250 \\
\bottomrule
\end{tabular*}
\end{table}

%% file: sections/results.tex
\input{tables/text_qa_main_results_error}

\section{Experiments}
\label{sec:results}

In this section, we verify our intuition about the components presented in equation~\eqref{eq:specific_bald_decomposition} and provide practical estimates across different domains, modalities, and tasks.

Specifically, we show the sensitivity of specification and residual aleatoric terms on several input modalities of LLMs - text (Section~\ref{sec:text_qa}), images (Section~\ref{sec:vqa}), and audio (Section~\ref{sec:audio_qa}).
Additionally, we evaluate the specificity of the SEA components in image classification, including out-of-distribution (OOD) detection for the epistemic term, in Appendix~\ref{app:image-classification}.

\textbf{Models.}
All LLM and VLM experiments used a heterogeneous ensemble of 4 models - Qwen2.5-Omni-7B~\citep{xu2025qwen25omni}, Gemma-4-E4B-it~\citep{gemmateam2026gemma4}, Phi-4-multimodal-instruct~\citep{abouelenin2025phi4mini}, and MiniCPM-o-4.5~\citep{cui2026minicpmo45}.
For text-to-text question answering (QA), we used a checkpoint ensemble, fine-tuned from Gemma-E2B-it and Gemma-E4B-it, on the AmbigQA~\citep{min2020ambigqa} dataset.
Full details on the fine-tuning procedure are presented in Appendix~\ref{app:ensemble_training_inference}.

\subsection{TextQA}
\label{sec:text_qa}

\textbf{Data.}
For text-to-text QA, we evaluate the specificity of SEA terms on two datasets:
\begin{enumerate}[leftmargin=15pt,labelindent=0pt]
    \item \textbf{CoQAR-U}. In the CoQAR dataset~\citep{brabant2022coqar}, the system is given a text passage and a question.
To model underspecified inputs, we remove the span with key information from the passage. Original passages serve as well-specified input.
We use GPT-5.6 Luna as $\pi(z | x)$ and sample five clarifications for each original and damaged passage (see details in Appendix~\ref{app:clarification_prompt_examples}).
    \item \textbf{AmbigQA}. We also use the existing AmbigQA~\citep{min2020ambigqa} dataset, which is based on an open-domain version of Natural Questions~\citep{kwiatkowski2019naturalquestions}.
Here, the system is given only a question, without any supporting context, and should answer based on its parametric memory. We treat questions with a single interpretation as well-specified, and questions with 2 or more interpretations as underspecified. Similar to CoQAR, we sample 5 clarifications per question using GPT-5.6 Luna.
\end{enumerate}

\textbf{Findings.}
First, Table \ref{tab:text_qa_main_results} shows that the specification component indeed performs well at detecting underspecified questions, aligning with its intended interpretation. 
Second, we check whether the residual aleatoric component in SEA decomposition still predicts in-domain error.
Table \ref{tab:text_qa_main_results_error} shows that the residual term delivers a higher ROC-AUC compared to other components.
We note that the magnitude of the residual term is often larger than that of the epistemic and specification terms; therefore, total uncertainty is mostly driven by the residual component, explaining the high performance of total uncertainty on this task.
%

\subsection{Visual QA}
\label{sec:vqa}

\input{tables/vqa_main_results}
\textbf{Data.}
We evaluate specification uncertainty in VQA by progressively hiding image content while keeping the question and reference answer fixed.
For each masked observation, five inpainted images provide alternative visual clarifications (see details in Appendix~\ref{app:vqa-experiments}). We used two VQA datasets:

\begin{enumerate}
    \item \textbf{VQS.} The frozen evaluation subset contains 500 questions on 500 distinct images sampled from the validation split of the VQS~\citep{gan2017vqs} dataset.
    Input underspecification is modeled by applying random gray masks covering 10\% and 30\% of the image area. Input clarifications were generated with FLUX.1 Fill-dev~\citep{blackforestlabs2024fluxfill}.
    Answer correctness is evaluated via Exact Match with the most common normalized human answer.
    \item \textbf{Imagenet.} The VQA dataset is built from ImageNet~\citep{deng2009imagenet} samples used in classification experiments by asking a general question, "What is depicted in the picture?" for every image.
    Masking and clarification construction follow the classification data construction pipeline (Appendix~\ref{app:image-masking-inpainting}).
    GPT-5.6 Luna evaluates the correctness of the answer by comparing the model's answer with the ImageNet class name.
\end{enumerate}

\textbf{Findings.}
Each SEA component is the strongest signal for its intended target. Specification uncertainty is the best underspecification detector on both datasets, while other components do not provide a useful signal for requesting a clarification (see Table~\ref{tab:vqa_main_results}).
Conversely, error detection is dominated by total and aleatoric uncertainty; since SEA preserves the total of the standard decomposition, isolating the specification term do not harm the error detection performance.
The Appendix~\ref{app:vqa-experiments} shows that both results hold for all six entropy estimators and all confidence baselines.

\begin{figure}[t]
    \centering
    \includegraphics[width=0.9\linewidth]{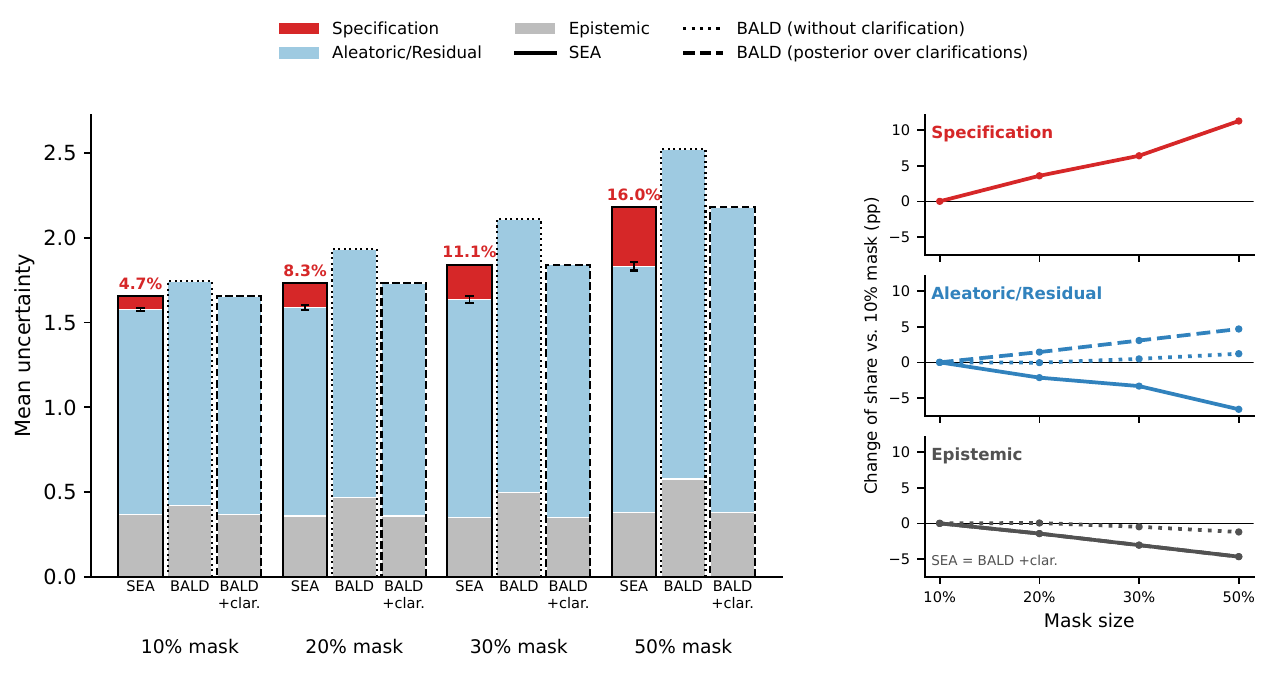}
    \caption{Uncertainty decomposition of the heterogeneous ensemble on ImageNet with the semantic entropy estimator. \textbf{Left:} mean uncertainty per mask size, stacked into epistemic, aleatoric, and specification components, for SEA, BALD on the masked image, and BALD with the posterior over clarifications. Red labels give the specification share of total uncertainty; error bars are 95\% intervals of the specification mean. \textbf{Right:} change of each component's share relative to the 10\% mask, in percentage points. Only the specification share grows with the masked area; SEA and BALD over clarifications have identical total and epistemic terms by construction.}
    \label{fig:vqa-decomposition-se}
\end{figure}

\textbf{Mask size effects.}
%
The ImageNet VQA setup allows control of the degree of underspecification in the input by varying the mask size relative to the object bounding box area.
Figure~\ref{fig:vqa-decomposition-se} compares how SEA and the two BALD variants split the ensemble's mean uncertainty on ImageNet as the masked fraction of the object bounding box grows.

Specification uncertainty grows monotonically with the amount of hidden content, both in absolute value and as a share of total uncertainty.
It is the only component whose share systematically increases. SEA and BALD over clarifications share the same total and epistemic terms by construction, since SEA only splits the pooled aleatoric term into residual aleatoric and specification uncertainty.
Under SEA, the aleatoric and epistemic shares decline with mask size, so the added uncertainty is absorbed almost entirely by the specification term, in agreement with its intended semantics.

Without the specification term, the missing information signal is hidden. BALD over clarifications folds the growth into its aleatoric term. BALD on the masked image inflates most with mask size, with both aleatoric and epistemic terms growing in absolute value while their 
shares stay nearly flat. Hence, its decomposition does not indicate that the extra uncertainty stems from missing input and could be removed by clarification rather than by changing the model.
%

\input{tables/audio_transposed_main_results}

\subsection{Audio QA}
\label{sec:audio_qa}
\textbf{Data.}
We render CoQAR-U passages as speech with OpenAI TTS, retaining the question as text.

\textbf{Findings}
Table~\ref{tab:audio_qa_main_results} shows that specification uncertainty is the strongest omission-detection signal, outperforming the runner-up total BALD uncertainty estimated on non-clarified input.
Thus, separating specification uncertainty recovers an underspecification-sensitive signal that is obscured in the total of the clarification-posterior.

%% file: tables/text_qa_main_results_error.tex
\begin{table}[t!]
\centering
\small
\setlength{\tabcolsep}{3pt}
\renewcommand{\arraystretch}{1.1}
\caption{Text QA error-detection AUROC using semantic entropy and GPT-5.6 Luna correctness labels. Errors in the ensemble answer to the original input are positive for every method. Evaluation uses 1,000 undamaged CoQAR-U questions and 830 unambiguous AmbigQA questions.}
\label{tab:text_qa_main_results_error}
\begin{tabular*}{\linewidth}{@{\extracolsep{\fill}}lcccccc@{}}
\toprule
 & \multicolumn{2}{c}{Gemma E2B} & \multicolumn{2}{c}{Gemma E4B} & \multicolumn{2}{c}{Heterogeneous} \\
\cmidrule(lr){2-3}\cmidrule(lr){4-5}\cmidrule(l){6-7}
Method & CoQAR-U & AmbigQA & CoQAR-U & AmbigQA & CoQAR-U & AmbigQA \\
\midrule
$\mathrm{SEA}_{\mathrm{res}}$   & 0.7285 & \textbf{0.8103} & 0.7535 & \textbf{0.7739} & 0.7077 & 0.7197 \\
$\mathrm{SEA}_{\mathrm{epi}}$   & 0.6573 & 0.6827 & 0.6145 & 0.6797 & 0.6730 & 0.5431 \\
$\mathrm{SEA}_{\mathrm{spec}}$  & 0.5627 & 0.5279 & 0.6600 & 0.5544 & 0.5456 & 0.7234 \\
$\mathrm{SEA}_{\mathrm{total}}$ & \textbf{0.7561} & 0.8062 & \textbf{0.7974} & 0.7710 & \textbf{0.7323} & \textbf{0.7412} \\
\midrule
\multicolumn{7}{@{}l}{BALD (without clarifications)} \\
$\mathrm{BALD}_{\mathrm{alea}}$  & 0.7665 & \textbf{0.8439} & 0.7800 & \textbf{0.8364} & 0.7438 & 0.7513 \\
$\mathrm{BALD}_{\mathrm{epi}}$   & 0.6628 & 0.6334 & 0.5326 & 0.6967 & 0.7214 & 0.5775 \\
$\mathrm{BALD}_{\mathrm{total}}$ & \textbf{0.7689} & 0.8435 & \textbf{0.7813} & \textbf{0.8364} & \textbf{0.7515} & \textbf{0.7692} \\
\midrule
\multicolumn{7}{@{}l}{BALD (clarification posterior)} \\
$\mathrm{BALD}_{\mathrm{alea}}$  & 0.7536 & \textbf{0.8062} & 0.7957 & 0.7709 & 0.7308 & 0.7267 \\
$\mathrm{BALD}_{\mathrm{epi}}$   & 0.6573 & 0.6827 & 0.6145 & 0.6797 & 0.6730 & 0.5431 \\
$\mathrm{BALD}_{\mathrm{total}}$ & \textbf{0.7561} & \textbf{0.8062} & \textbf{0.7974} & \textbf{0.7710} & \textbf{0.7323} & \textbf{0.7412} \\
\bottomrule
\end{tabular*}
\end{table}

%% file: tables/vqa_main_results.tex
\begin{table}[]
    \centering
    \small
    \setlength{\tabcolsep}{3pt}
    \renewcommand{\arraystretch}{1.1}
    \caption{VQA AUROC of the SEA components on VQS and ImageNet, computed with the semantic entropy estimator for the heterogeneous ensemble. Underspecification detection takes 30\% masking as the positive class and 10\% as the negative class. Error detection uses an incorrect ensemble answer to the masked image as the positive class, reported separately at each mask size.}
    \label{tab:vqa_main_results}
    \begin{tabular}{lcccccc}
\toprule
 & \multicolumn{2}{c}{Underspecification} & \multicolumn{4}{c}{ID error} \\
\cmidrule(lr){2-3} \cmidrule(lr){4-7}
 & VQS & ImageNet & \multicolumn{2}{c}{VQS} & \multicolumn{2}{c}{ImageNet} \\
\cmidrule(lr){2-2} \cmidrule(lr){3-3} \cmidrule(lr){4-5} \cmidrule(lr){6-7}
Method & 30\% vs 10\% & 30\% vs 10\% & 10\% & 30\% & 10\% & 30\% \\
\midrule
$\mathrm{SEA}_{\mathrm{residual}}$ & 0.5422 & 0.5289 & 0.8217 & 0.7850 & 0.7448 & 0.7715 \\
$\mathrm{SEA}_{\mathrm{epi}}$ & 0.5073 & 0.4907 & 0.7830 & 0.7521 & 0.7275 & 0.7334 \\
$\mathrm{SEA}_{\mathrm{spec}}$ & \textbf{0.6614} & \textbf{0.6903} & 0.7256 & 0.7006 & 0.5992 & 0.7172 \\
$\mathrm{SEA}_{\mathrm{total}}$ & 0.5486 & 0.5452 & \textbf{0.8289} & \textbf{0.7900} & \textbf{0.7458} & \textbf{0.7756} \\
\midrule
\multicolumn{7}{l}{BALD (without clarifications)} \\
$\mathrm{BALD}_{\mathrm{alea}}$ & \textbf{0.5363} & 0.5768 & 0.8194 & \textbf{0.8087} & 0.7151 & 0.7653 \\
$\mathrm{BALD}_{\mathrm{epi}}$ & 0.5223 & 0.5704 & 0.7968 & 0.7440 & 0.7193 & 0.7264 \\
$\mathrm{BALD}_{\mathrm{total}}$ & 0.5332 & \textbf{0.5806} & \textbf{0.8311} & 0.8019 & \textbf{0.7263} & \textbf{0.7742} \\
\midrule
\multicolumn{7}{l}{BALD (clarification posterior)} \\
$\mathrm{BALD}_{\mathrm{alea}}$ & \textbf{0.5618} & \textbf{0.5578} & 0.8268 & 0.7871 & 0.7393 & 0.7753 \\
$\mathrm{BALD}_{\mathrm{epi}}$ & 0.5073 & 0.4907 & 0.7830 & 0.7521 & 0.7275 & 0.7334 \\
$\mathrm{BALD}_{\mathrm{total}}$ & 0.5486 & 0.5452 & \textbf{0.8289} & \textbf{0.7900} & \textbf{0.7458} & \textbf{0.7756} \\
\bottomrule
\end{tabular}

\end{table}

%% file: tables/audio_transposed_main_results.tex

\begin{table}[t]
\centering
\small
\setlength{\tabcolsep}{6pt}
\renewcommand{\arraystretch}{1.1}
\caption{Audio QA underspecification-detection AUROC on CoQAR-U-TTS.}
\label{tab:audio_qa_main_results}
\begin{tabular}{@{}lccc@{}}
\toprule
 & SEA & BALD (without clarification) & BALD (clarification posterior) \\
\midrule
Total         & 0.5926 & \textbf{0.7427} & \textbf{0.5926} \\
Aleatoric     & 0.5170 & 0.7229 & 0.5851 \\
Epistemic     & 0.4937 & 0.6108 & 0.4937 \\
Specification & \textbf{0.7739} & -- & -- \\
\bottomrule
\end{tabular}
\end{table}

%% file: sections/conclusion.tex

\section{Conclusion}
\label{sec:conclusion}
In this paper, we considered a predictive uncertainty setting in which the observed input~\(x\) may only partially describe the intent.
We derived SEA, a decomposition of prediction risk into specification, residual aleatoric, and epistemic uncertainty.
At the original information level, the first two terms split the Bayes risk into the expected gain from clarification and the remaining risk, and each term corresponds to an action.
In QA over text, images, and audio, specification uncertainty is the most informative score for detecting underspecified inputs, even where uncertainty computed only on~\(x\), without clarifications, fails to detect them.
Residual aleatoric uncertainty is the most informative component for ID error detection in all but one setting. The standard decomposition, even when sensitive to underspecification, cannot show whether clarification would reduce the risk.

\subsection*{AI use statement}

In this work, we used generative AI tools for generating clarifications for SEA estimation in all datasets, implementing the method and experiments, reformatting AmbigQA dataset, preparing visualization code, polishing text in both main and appendix sections and drafting some appendix sections. We have not used generative AI tools for generating research hypotheses, theory derivation, proofs, experiment design or quality metric choice. We take responsibility for the final content of this work, including text, claims or artifacts produced with the aid of generative AI.

\subsection*{Reproducibility statement}

For theoretical results, complete
proof of the claims is presented in Section \ref{sec:method} and appendices \ref{app:other_approximations}, \ref{app:bayes-risk-split}.

For experimental results, we elaborate the details of practical estimation of SEA decomposition in appendix \ref{app:mc_estimates}, details of LLM ensemble training in appendix \ref{app:ensemble_training_inference}, and describe the process of clarification generation in appendix \ref{app:clarification_prompt_examples}.

We will also release all custom datasets (e.g. CoQAR-U, CoQAR-U-TTS, AmbigQA along with sampled clarifications) and experimental pipelines upon acceptance.

%% file: sections/appendix.tex

\section{Other approximation strategies}
\label{app:other_approximations}

In Section~\ref{sec:method}, we estimate the terms of equation~\eqref{eq:gt_risk_decomposition} with approximation \((1,2)\) of \citet{kotelevskii2025risk}.
Here, we describe the approximation ideas they consider, apply them to equation~\eqref{eq:gt_risk_decomposition}, and give the resulting estimates for several proper scoring rules.

\paragraph{Notation.}
We use \(q_{z,\theta}\), \(q_\theta\), and \(q\) from Section~\ref{sec:method} and additionally define the posterior predictive distribution given the intent,
\begin{equation*}
    \bar q_z \coloneqq \EE_\pi[q_{Z,\Theta}\mid Z=z] = q(y\mid z,x,D).
\end{equation*}
Expectations are taken over all random variables inside them, where \((Z,\Theta)\) is distributed according to \(\pi(\theta,z\mid x,D)\).
When an estimate uses two posterior samples of the parameters, we denote them by \(\tilde\Theta\) and \(\Theta\), which are independent and identically distributed.
Under the factorization \(\pi(\theta,z\mid x,D)=\pi(\theta\mid D)\pi(z\mid x)\) used in our experiments, \(\bar q_z = \EE_{\Theta}[q_{z,\Theta}]\).

\paragraph{Approximation ideas.}
\citet{kotelevskii2025risk} view a risk as a function \(g(p,q)\) of the ground-truth distribution \(p\) and the predictor \(q\).
Given posterior samples \(r_\Theta\) of a distribution, they replace the corresponding argument of \(g\) in one of three ways:
\begin{enumerate}
    \item \emph{Bayesian averaging of risk.} Evaluate \(g\) at each sample \(r_\Theta\) and average the result over the posterior.
    \item \emph{Central label.} Plug in the posterior predictive distribution \(\operatorname*{arg\,min}_r\EE[D_G(r_\Theta~\|~r)]=\EE[r_\Theta]\).
    \item \emph{Central prediction.} Plug in \(\operatorname*{arg\,min}_r\EE[D_G(r~\|~r_\Theta)]\).
\end{enumerate}
Approximation \((i,j)\) applies idea \(i\) to the ground-truth argument and idea \(j\) to the predictor.
For brevity, we consider only ideas 1 and 2 below, which give four approximations \((i,j)\) with \(i,j\in\{1,2\}\).

\paragraph{Application to our decomposition.}
In our setting, the ground-truth argument is the whole process \(p(y,z\mid x)\).
Idea 1 replaces it by the process \(q(y\mid z,\theta)\pi(z\mid x,\theta,D)\) of each posterior sample \(\theta\), with outcome distributions \(q_{z,\theta}\) and marginal \(q_\theta\).
Idea 2 replaces it by the posterior predictive process \(q(y,z\mid x,D)\), with outcome distributions \(\bar q_z\) and marginal \(q\).
Here, \(\bar q_z\) and \(q\) are the central labels of the samples \(q_{z,\Theta}\) given \(Z=z\) and of the samples \(q_\Theta\), respectively.
The predictor does not observe the intent, so the two ideas use \(q_\theta\) for each posterior sample and the posterior predictive distribution \(q\), respectively.

\paragraph{Resulting estimates.}
The residual aleatoric and specification terms depend only on the process, so each has two estimates,
\begin{align*}
    \widehat R_{\mathrm{res}}^{(1)} &= -\EE[G(q_{Z,\Theta})], &
    \widehat R_{\mathrm{spec}}^{(1)} &= \EE[D_G(q_{Z,\Theta}~\|~q_\Theta)],\\
    \widehat R_{\mathrm{res}}^{(2)} &= -\EE[G(\bar q_Z)], &
    \widehat R_{\mathrm{spec}}^{(2)} &= \EE[D_G(\bar q_Z~\|~q)].
\end{align*}
Their sums are the corresponding estimates of the Bayes risk of \citet{kotelevskii2025risk}, \(\widehat R_b^{(1)} = -\EE[G(q_\Theta)]\) and \(\widehat R_b^{(2)} = -G(q)\).
The epistemic term depends on both arguments, which gives four estimates,
\begin{align*}
    \widehat R_{\mathrm{exc}}^{(1,1)} &= \EE[D_G(q_{\tilde\Theta}~\|~q_\Theta)], &
    \widehat R_{\mathrm{exc}}^{(1,2)} &= \EE[D_G(q_\Theta~\|~q)],\\
    \widehat R_{\mathrm{exc}}^{(2,1)} &= \EE[D_G(q~\|~q_\Theta)], &
    \widehat R_{\mathrm{exc}}^{(2,2)} &= D_G(q~\|~q) = 0.
\end{align*}
\citet{kotelevskii2025risk} call \(\widehat R_{\mathrm{exc}}^{(1,1)}\), \(\widehat R_{\mathrm{exc}}^{(1,2)}\), and \(\widehat R_{\mathrm{exc}}^{(2,1)}\) the expected pairwise Bregman divergence (EPBD), the Bregman information (BI), and the reverse Bregman information (RBI).
The total risk \(R(p,q)\) is linear in \(p\), so both ideas for the process give the same estimate \(\widehat R^{(\cdot,j)}\coloneqq\widehat R^{(1,j)}=\widehat R^{(2,j)}\) with
\begin{equation*}
    \widehat R^{(\cdot,1)} = \EE[R(q,q_\Theta)] = -G(q)+\EE[D_G(q~\|~q_\Theta)], \qquad
    \widehat R^{(\cdot,2)} = R(q,q) = -G(q).
\end{equation*}

\paragraph{Decompositions.}
Applying equation~\eqref{eq:gt_risk_decomposition} to the estimated process and predictor, and averaging over the posterior samples for idea 1, we obtain for every approximation \((i,j)\)
\begin{equation*}
    \widehat R^{(\cdot,j)} = \widehat R_{\mathrm{res}}^{(i)} + \widehat R_{\mathrm{spec}}^{(i)} + \widehat R_{\mathrm{exc}}^{(i,j)}.
\end{equation*}
For \((i,j)=(1,2)\), this is equation~\eqref{eq:our_decomposition_general}.
Approximation \((2,2)\), by construction, estimates epistemic uncertainty as zero. It is equation~\eqref{eq:single_model_general_decomposition} with \(q_z\) replaced by \(\bar q_z\), i.e., it applies the decompositions of \citet{input_clarification} and \citet{walha2026finegraineduncertaintydecompositioninlargelanguagemodelsaspectralapproach} to the posterior predictive model.
For a single model, \(\bar q_z = q_z\), and all four approximations reduce to equation~\eqref{eq:single_model_general_decomposition}.

\paragraph{Examples.}
Table~\ref{tab:approximation_examples} lists the estimates for three proper scoring rules.

\emph{Log score.}
Here \(G(r)=\sum_y r(y)\log r(y)\), so \(-G(r)=\HH(r)\), \(D_G\) is the KL divergence, and \(R(p,r)=\mathrm{CE}(p~\|~r)\coloneqq-\sum_y p(y)\log r(y)\) is the cross-entropy.
Under the joint distribution \(q(y,z,\theta\mid x,D)\), \(q_{z,\theta}\) is the distribution of \(Y\) given \((Z,\Theta)=(z,\theta)\), \(\bar q_z\) is its distribution given \(Z=z\), \(q_\theta\) given \(\Theta=\theta\), and \(q\) is its marginal distribution.
Hence, the estimates become entropies and mutual informations.
In particular, approximation \((1,2)\) gives the chain rule \(\HH(Y)=\HH(Y\mid Z,\Theta)+I(Y;Z\mid\Theta)+I(Y;\Theta)\), i.e., equation~\eqref{eq:specific_bald_decomposition}, and approximation \((2,2)\) gives \(\HH(Y)=\HH(Y\mid Z)+I(Y;Z)\).
The epistemic estimates of approximations \((1,1)\) and \((2,1)\) are the expected pairwise KL divergence (EPKL) and the reverse mutual information (RMI)~\citep{malinin2021uncertainty}.

\emph{Brier score.}
Here \(G(r)=\|r\|_2^2-1\), where \(r\) is viewed as a vector in \(\RR^{|\Y|}\).
Then \(\ell(r,y)=\|r-e_y\|_2^2\), where \(e_y\) is the one-hot vector of \(y\), and \(D_G(r~\|~s)=\|r-s\|_2^2\).
Hence, the specification and epistemic estimates are variances of the predictions, summed over the classes.
For example, \(\widehat R_{\mathrm{spec}}^{(1)}\) is the expected variance of \(q_{Z,\Theta}\) across intents, and \(\widehat R_{\mathrm{exc}}^{(1,2)}\) is the variance of \(q_\Theta\) across posterior samples.
Since \(D_G\) is symmetric, \(\widehat R_{\mathrm{exc}}^{(2,1)}=\widehat R_{\mathrm{exc}}^{(1,2)}\), and since \(\tilde\Theta\) and \(\Theta\) are independent, \(\widehat R_{\mathrm{exc}}^{(1,1)}=2\widehat R_{\mathrm{exc}}^{(1,2)}\).

\emph{Zero-one score.}
Here \(G(r)=M(r)-1\), where \(M(r)\coloneqq\max_y r(y)\).
With the subgradient \(G'(r)=e_{\hat y(r)}\), where \(\hat y(r)\in\arg\max_y r(y)\) is the predicted class, \(\ell(r,y)=\mathbf{1}\{y\neq\hat y(r)\}\) is the zero-one loss and \(D_G(r~\|~s)=M(r)-r(\hat y(s))\).
The estimates \(\widehat R_{\mathrm{spec}}^{(1)}\), \(\widehat R_{\mathrm{spec}}^{(2)}\), and \(\widehat R_{\mathrm{exc}}^{(1,2)}\) are Bregman informations, since the second argument of \(D_G\) is the (conditional) mean of the first. Hence, they reduce to differences of expected maximal probabilities.
In particular, \(\widehat R_{\mathrm{spec}}^{(1)}\) is the expected increase in the probability of a correct prediction once the intent is revealed, computed under the process of each posterior sample.
This score is proper but not strictly proper.

\begin{table}[ht]
\centering
\caption{Estimates of the terms of equation~\eqref{eq:gt_risk_decomposition} for three proper scoring rules. For the log score, the information-theoretic quantities are computed under \(q(y,z,\theta\mid x,D)\), and we omit the conditioning on \(X=x\) and \(D\). Here, \(M(r)=\max_y r(y)\) and \(\hat y_\Theta=\hat y(q_\Theta)\).}
\label{tab:approximation_examples}
\small
\renewcommand{\arraystretch}{1.2}
\begin{tabular}{@{}llll@{}}
\toprule
Estimate & Log score & Brier score & Zero-one score \\
\midrule
\(\widehat R_{\mathrm{res}}^{(1)}\) & \(\HH(Y\mid Z,\Theta)\) & \(1-\EE\|q_{Z,\Theta}\|_2^2\) & \(1-\EE[M(q_{Z,\Theta})]\) \\
\(\widehat R_{\mathrm{res}}^{(2)}\) & \(\HH(Y\mid Z)\) & \(1-\EE\|\bar q_Z\|_2^2\) & \(1-\EE[M(\bar q_Z)]\) \\
\(\widehat R_{\mathrm{spec}}^{(1)}\) & \(I(Y;Z\mid\Theta)\) & \(\EE\|q_{Z,\Theta}-q_\Theta\|_2^2\) & \(\EE[M(q_{Z,\Theta})]-\EE[M(q_\Theta)]\) \\
\(\widehat R_{\mathrm{spec}}^{(2)}\) & \(I(Y;Z)\) & \(\EE\|\bar q_Z-q\|_2^2\) & \(\EE[M(\bar q_Z)]-M(q)\) \\
\midrule
\(\widehat R_{\mathrm{exc}}^{(1,1)}\) & \(\EE[KL(q_{\tilde\Theta}~\|~q_\Theta)]\) & \(2\,\EE\|q_\Theta-q\|_2^2\) & \(\EE[M(q_\Theta)-q(\hat y_\Theta)]\) \\
\(\widehat R_{\mathrm{exc}}^{(1,2)}\) & \(I(Y;\Theta)\) & \(\EE\|q_\Theta-q\|_2^2\) & \(\EE[M(q_\Theta)]-M(q)\) \\
\(\widehat R_{\mathrm{exc}}^{(2,1)}\) & \(\EE[KL(q~\|~q_\Theta)]\) & \(\EE\|q_\Theta-q\|_2^2\) & \(M(q)-\EE[q(\hat y_\Theta)]\) \\
\midrule
\(\widehat R^{(\cdot,1)}\) & \(\EE[\mathrm{CE}(q~\|~q_\Theta)]\) & \(1-\|q\|_2^2+\EE\|q_\Theta-q\|_2^2\) & \(1-\EE[q(\hat y_\Theta)]\) \\
\(\widehat R^{(\cdot,2)}\) & \(\HH(Y)\) & \(1-\|q\|_2^2\) & \(1-M(q)\) \\
\bottomrule
\end{tabular}
\end{table}

\paragraph{Relations between the estimates.}
As shown by \citet{kotelevskii2025risk}, \(\widehat R_{\mathrm{exc}}^{(1,1)}=\widehat R_{\mathrm{exc}}^{(1,2)}+\widehat R_{\mathrm{exc}}^{(2,1)}\), and the two estimates of the Bayes risk differ by the Bregman information, \(\widehat R_b^{(2)}-\widehat R_b^{(1)}=\widehat R_{\mathrm{exc}}^{(1,2)}\geq 0\).
Similarly, since \(\EE[q_{Z,\Theta}\mid Z]=\bar q_Z\), the two estimates of the residual aleatoric term satisfy \(\widehat R_{\mathrm{res}}^{(2)}-\widehat R_{\mathrm{res}}^{(1)}=\EE[D_G(q_{Z,\Theta}~\|~\bar q_Z)]\geq 0\).
Hence, the two estimates of the specification term differ by \(\widehat R_{\mathrm{spec}}^{(1)}-\widehat R_{\mathrm{spec}}^{(2)}=\EE[D_G(q_{Z,\Theta}~\|~\bar q_Z)]-\widehat R_{\mathrm{exc}}^{(1,2)}\), which can have either sign.
If \(Z\) and \(\Theta\) are independent under \(\pi\), as in our experiments, this difference is nonnegative for the log and Brier scores, where it equals \(I(Z;\Theta\mid Y)\) and \(\EE\|q_{Z,\Theta}-q_\Theta-\bar q_Z+q\|_2^2\), respectively.

\section{Bayes risk split}
\label{app:bayes-risk-split}

We derive the split of the Bayes risk used in equation~\eqref{eq:gt_risk_decomposition}.
Below, as in the main part, we assume that \(x\) is fixed.

By the definition of the Bregman divergence,
\begin{equation*}
    \EE_Z[D_G(p_Z~\|~p)]
    = \EE_Z[G(p_Z)] - G(p) - \scalarprod{G'(p)}{\EE_Z[p_Z] - p}
    = \EE_Z[G(p_Z)] - G(p),
\end{equation*}
where the last equality uses \(\EE_Z[p_Z] = p\).

Rearranging the terms gives
\begin{equation*}
    -G(p) = -\EE_Z[G(p_Z)] + \EE_Z[D_G(p_Z~\|~p)],
\end{equation*}
and substituting this into equation~\eqref{eq:proper-scoring-rules-decomposition-risk} yields equation~\eqref{eq:gt_risk_decomposition}.

Interestingly, we can show that since \(q\) does not depend on \(z\), we have \(R(p,q) = \EE_{Z\sim p(\cdot\mid x)}\big[\EE_{Y\sim p_Z}[\ell(q,Y)]\big] = \EE_{Z\sim p(\cdot\mid x)}[R(p_Z,q)]\), i.e., \(R(p,q)\) is also the risk of \(q\) under the \emph{joint process} \(p(y, z \mid x)\). This holds for any loss that does not depend on \(z\).

\section{Limitations}
\label{app:limitations}

We see two limitations of our approach.

\textbf{Approximation of the intent distribution.}
Recall that our approach is built on separating specification uncertainty from residual aleatoric uncertainty. This requires the distribution \( p(z \mid x) \) over possible values of the unknown intent.
In the Bayesian model of Section~\ref{sec:bayesian-approximations}, intents and parameters have a joint distribution \( \pi(\theta, z \mid x, \mathcal{D}) \), so the intent distribution in general may depend on the training data and on \( \theta \).
In our experiments, we instead use the factorization \( \pi(\theta, z \mid x, \mathcal{D}) = \pi(\theta \mid \mathcal{D})\pi(z \mid x) \), where \( \pi(z \mid x) \) is given by external generative models and depends on neither \( \mathcal{D} \) nor \( \theta \).
The estimates, therefore, inherit the generators' view of which intents are plausible. A generator whose intents are less diverse than the true ones can underestimate specification uncertainty, and one that proposes implausible intents can overestimate it.

This dependence is, however, inherent to the problem rather than specific to our approach.
As discussed in Section~\ref{sec:bayesian-approximations}, pairs \( (x, y) \) determine \( p(y \mid x) \) but not how its Bayes risk splits into specification and residual aleatoric uncertainty.
Any estimate of this split, including those of \citet{input_clarification} and \citet{walha2026finegraineduncertaintydecompositioninlargelanguagemodelsaspectralapproach}, therefore requires a model of the intents, for which large pretrained generative models are currently the most practical choice.
Empirically, the specification term was the strongest signal of underspecification across each clarification pipeline we used, from LLM-generated text to direct inpainting with a fixed prompt and no language model (Section~\ref{sec:results} and Appendix~\ref{app:vqa-experiments}).
This suggests that the signal is not tied to a particular generator.

\textbf{Estimation of epistemic uncertainty.}
The epistemic term \( D_G(p \,\|\, q) \) cannot be estimated from the predictive distribution \( q \) alone.
It requires either a second-order model, such as an ensemble whose members serve as samples from \( \pi(\theta \mid \mathcal{D}) \), or an auxiliary model that directly estimates the excess risk \citep{kotelevskii2022nonparametric, lahlou2021deup}.
Both options add cost. Ensembles multiply the training and inference cost by the number of members, which is especially restrictive for LLMs, while auxiliary models have to be trained separately.
This requirement is not specific to SEA, since the epistemic term of the standard information-theoretic decomposition also requires a second-order model.

\section{Monte-Carlo Estimates of Components}
\label{app:mc_estimates}
For fixed \(x\) and training data \(D\), we use the predictive distributions from Section~\ref{sec:method}:
\begin{equation}
    q_{z,\theta} \coloneqq q(y\mid z,\theta).
\end{equation}
As in our experiments, we assume \(\pi(\theta,z\mid x,D)=\pi(\theta\mid D)\pi(z\mid x)\).
Let \(\theta_m\sim\pi(\theta\mid D)\), \(m=1,\ldots,M\), denote the parameter samples represented by the ensemble, and let \(z_k\sim\pi(z\mid x)\), \(k=1,\ldots,K\), denote the sampled intents shared across its members.
Here, \(M\) is the ensemble size and \(K\) is the number of intents.

The clarification-marginalized predictive distribution of one member and the posterior predictive distribution are approximated by
\begin{equation}
    q_\theta
    = \EE_\pi[q_{Z,\Theta}\mid\Theta=\theta]
    \approx \widehat q_\theta
    \coloneqq \frac{1}{K}\sum_{k=1}^K q_{z_k,\theta},
\end{equation}
\begin{equation}
    \begin{aligned}
        q &= \EE_\pi[q_{Z,\Theta}] = q(y\mid x,D) \\
          &\approx \widehat q
          \coloneqq \frac{1}{M}\sum_{m=1}^M \widehat q_{\theta_m}
          = \frac{1}{MK}\sum_{m=1}^M\sum_{k=1}^K q_{z_k,\theta_m}.
    \end{aligned}
\end{equation}
Hats distinguish finite-sample estimates from the corresponding distributions and uncertainty components.
All expectations below are taken under \(\pi(\theta,z\mid x,D)\), as in equation~\eqref{eq:specific_bald_decomposition}.
Its three components have the following plug-in Monte Carlo estimates.

\paragraph{Residual aleatoric.}
\begin{equation}
    \begin{aligned}
        \mathrm{SEA}_{\mathrm{res}}
        &= \EE_{Z,\Theta}[\HH(q_{Z,\Theta})] \\
        &\approx \widehat{\mathrm{SEA}}_{\mathrm{res}}
        \coloneqq \frac{1}{MK}\sum_{m=1}^M\sum_{k=1}^K
        \HH(q_{z_k,\theta_m}).
    \end{aligned}
\end{equation}

\paragraph{Specification.}
\begin{equation}
    \begin{aligned}
        \mathrm{SEA}_{\mathrm{spec}}
        &= \EE_{Z,\Theta}[KL(q_{Z,\Theta}~\|~q_\Theta)] \\
        &= \EE_\Theta[\HH(q_\Theta)]-\mathrm{SEA}_{\mathrm{res}} \\
        &\approx \widehat{\mathrm{SEA}}_{\mathrm{spec}}
        \coloneqq \frac{1}{M}\sum_{m=1}^M\HH(\widehat q_{\theta_m})
        -\widehat{\mathrm{SEA}}_{\mathrm{res}}.
    \end{aligned}
\end{equation}

\paragraph{Epistemic.}
\begin{equation}
    \begin{aligned}
        \mathrm{SEA}_{\mathrm{epi}}
        &= \EE_\Theta[KL(q_\Theta~\|~q)]
        = \HH(q)-\EE_\Theta[\HH(q_\Theta)] \\
        &\approx \widehat{\mathrm{SEA}}_{\mathrm{epi}}
        \coloneqq \HH(\widehat q)
        -\frac{1}{M}\sum_{m=1}^M\HH(\widehat q_{\theta_m}).
    \end{aligned}
\end{equation}
Using the same sampled distributions in all three terms preserves the decomposition:
\begin{equation*}
    \widehat{\mathrm{SEA}}_{\mathrm{total}}
    \coloneqq \HH(\widehat q)
    = \widehat{\mathrm{SEA}}_{\mathrm{res}}
    + \widehat{\mathrm{SEA}}_{\mathrm{spec}}
    + \widehat{\mathrm{SEA}}_{\mathrm{epi}}.
\end{equation*}

\input{sections/llm_ensembling}
\input{sections/semantic_entropy}

\section{Ensemble training and inference details.}
\label{app:ensemble_training_inference}

\paragraph{Ensemble details.} 
Since training an ensemble of language models from scratch is prohibitively expensive, we turn to fine-tuning. We tune 5 Gemma-4-E4B-it and Gemma-4-E2B-it models~\citep{gemmateam2026gemma4} for short-form factual question answering on the AmbigQA dataset~\citep{min2020ambigqa}. All models are trained on the same training set, differing only in the order of examples.
We tried LoRA and Orthogonal finetuning (OFT) \citep{hu2022lora, qiu2025oft}. Both configurations yielded similar AmbigQA validation performance, but OFT showed less catastrophic forgetting on MMLU and IFEval subsets, so we report results for OFT ensembles in the main text.


\paragraph{Fine-tuning data.} Original AmbigQA contains questions of two types: well-formed, admitting a single answer, and ambiguous, which have multiple interpretations. For each ambiguous question, the dataset contains from 2 to 9 clarified versions of this question along with the corresponding ground truth answer. For fine-tuning, we only use well-formed or clarified questions, not ambiguous ones, since it's unclear what should be a short-form ground truth answer for an ambiguous question. 

We split official train set of AmbigQA into local train and development set in 90/10 proportion. Our final training set contains 17261 examples, and the development set contains 1982 examples.

\paragraph{Ensemble training regime}
Each model is fine-tuned for three epochs using OFT with a block size of 32. We target all linear layers, apply a module dropout rate of 0.05, and use the Cayley--Neumann approximation with five terms; biases are not adapted. Training uses a per-device batch size of 16 with two gradient-accumulation steps, corresponding to an effective batch size of 32 per device. We set the maximum input length to 512 tokens and optimize with a learning rate of $5\times10^{-5}$, weight decay of 0.01, and a linear warm-up over the first 3\% of training steps. Gradients are clipped to a maximum norm of 1.0. Computation uses bfloat16 precision with TF32 enabled, and evaluation is performed every 150 optimization steps. We use the AdamW optimizer.

\section{Performance of the ensembles}
\label{app:text-performance}
We evaluate answer correctness with GPT-5.6 Luna~\citep{openai2026gpt56luna}, using temperature 0, disabled reasoning, and a maximum of 32 output tokens.
The judge receives the original question, dataset reference answer(s), and selected ensemble answer; for CoQAR-U, it also receives the original undamaged passage.
It returns 1 if the answer addresses the question and agrees in meaning with at least one reference (and with the passage for CoQAR-U), and 0 otherwise, accepting paraphrases and harmless extra detail but rejecting contradictions and nonanswers.
We select the highest-probability candidate from the saved ensemble answer pool; for the clarification posterior, selection follows averaging probabilities over ensemble members and five clarifications.
Table~\ref{tab:text_ensemble_performance} reports the mean binary judgment in each input regime, counting each source question once per regime.

Unsurprisingly, omissions and ambiguity reduce performance, and sampling possible clarifications restores it only somewhat. This highlights the need to first detect and then resolve ambiguities and a lack of specification by asking the user additional questions with the goal of restoring the true original intent.

\input{tables/text_ensemble_performance}

\FloatBarrier
\input{sections/text_correctness_prompts}

\section{Ensemble-free estimation}
\label{app:ensemble_free_llm}
Tables~\ref{tab:text_ensemble_free_underspec} and ~\ref{tab:vqa_ensemble_free_underspec} report underspecification detection for each member of the heterogeneous ensemble evaluated separately on text  and visual QA. We restrict the answer support and sequence probabilities to the selected model, retaining its ten draws per clarification and the saved semantic cluster assignments. With one model, the epistemic component vanishes, while specification uncertainty still measures variation across clarifications. The specification term achieves the highest AUROC for every model on both text datasets. Thus, if one is willing to sacrifice the epistemic component, efficient underspecification detection can still be performed with a single model.

\input{tables/text_ensemble_free_underspec}

\input{tables/vqa_ens_free_underspec}





\clearpage
\input{sections/clarification_generation}
\input{sections/image_domain}

\input{sections/audio_domain}

\input{sections/vqa_experiments}

%% file: sections/llm_ensembling.tex
\section{LLM Ensembling}
\label{app:llm-ensembling}

We implement the predictive mixtures in Appendix~\ref{app:mc_estimates} by separating answer sampling from probability evaluation. The same set of clarifications is shared across all ensemble members, so variation across clarifications and variation across models remain distinct axes.

\paragraph{Separate sampling and a shared answer pool.}
For each member $\theta_m$ and clarification $z_k$, we sample $S$ answers from $q_{z_k,\theta_m}$; our text QA experiments use $K=5$ clarifications and $S=10$ answers per pair. Different pairs can produce different, even disjoint, answer sets. An answer missing from one pair's samples may nevertheless have substantial probability under that pair, so treating unobserved answers as having zero probability would distort the mixture and its uncertainty. We therefore form a shared candidate set $\mathcal U$ from the union of all $MKS$ draws. We deduplicate candidates for scoring while retaining every draw's originating model, clarification, and multiplicity for Monte Carlo estimation.

\paragraph{Cross-scoring.}
Every member scores every candidate $u\in\mathcal U$ under every clarification, including candidates produced only by other members or clarifications. With teacher forcing, the candidate is supplied as the answer continuation, and we sum the conditional log-probabilities of its tokens:
\begin{equation}
    \ell_{mku}
    = \sum_{t=1}^{T_m(u)}
      \log q\bigl(u_t^{(m)}\mid u_{<t}^{(m)},z_k,\theta_m\bigr),
    \label{eq:llm-cross-scoring}
\end{equation}
where $u^{(m)}$ is the candidate's tokenization for member $m$, and prompt tokens do not contribute to the sum. This produces a dense $M\times K\times|\mathcal U|$ score tensor despite potentially disjoint sampled answer sets. For checkpoint ensembles with a shared tokenizer, candidate identity is determined by token sequences after the configured termination-token handling. For heterogeneous ensembles, we pool cleaned answer strings and tokenize each string with the scoring member's own tokenizer. All mixture scores, including those for a candidate's generating pair, come from this common cross-scoring procedure; generation-time scores are retained only as diagnostics.

\paragraph{Averaging over clarifications and models.}
We average \emph{probabilities}, first over clarifications within a member and then over members:
\begin{equation}
    \widehat q_{\theta_m}(u)
    = \frac{1}{K}\sum_{k=1}^{K}\exp(\ell_{mku}),
    \qquad
    \widehat q(u)
    = \frac{1}{M}\sum_{m=1}^{M}\widehat q_{\theta_m}(u).
    \label{eq:llm-predictive-mixtures}
\end{equation}
These averages are evaluated in log space using log-sum-exp. The individual, clarification-marginalized, and fully averaged scores provide the distributions needed for the SEA components and clarification-posterior BALD. For answer-level correctness evaluation, we select $\arg\max_{u\in\mathcal U}\widehat q(u)$. BALD without clarifications uses a separate answer pool sampled and cross-scored on the original input $x$.

%% file: sections/semantic_entropy.tex
\begin{samepage}
\section{Semantic Entropy Implementation}
\label{app:semantic-entropy}

We cluster the pooled answers $\mathcal U$ once, giving all model--clarification pairs the same answer-to-class mapping $g$. For a component or mixture predictive distribution $r$, we approximate each class's mass and estimate semantic entropy as
\begin{equation}
    \widehat P_r(c)
    = \sum_{u\in\mathcal U:\,g(u)=c} r(u),
    \qquad
    \widehat H_{\mathrm{SE}}(r)
    = -\frac{1}{n_r}\sum_{i=1}^{n_r}
      \log\widehat P_r\bigl(g(y_i)\bigr).
    \label{eq:semantic-entropy-implementation}
\end{equation}
\end{samepage}
Here, $r(u)$ is obtained by cross-scoring as in Appendix~\ref{app:llm-ensembling}, and $y_i$ are the matched Monte Carlo draws: $S$ draws for a model--clarification pair, $KS$ draws pooled over clarifications for a member, or $MKS$ draws for the full mixture. Repeated draws retain their multiplicity. Class masses are evaluated on the finite candidate pool without renormalization or sequence-length normalization. The resulting entropy estimates are combined as in Appendix~\ref{app:mc_estimates}.

For natural language inference (NLI), we use \texttt{microsoft/deberta-large-mnli}. An answer joins an existing class when the NLI model predicts entailment in both directions between that answer and the class representative; otherwise, it starts a new class. In the AmbigQA E2B and heterogeneous runs, the complete encoded NLI pair is capped at 100 tokens, including special tokens, using longest-first truncation across the two answers.

%% file: tables/text_ensemble_performance.tex
\begin{table}[htbp]
\centering
\small
\setlength{\tabcolsep}{3pt}
\renewcommand{\arraystretch}{1.1}
\caption{Text QA accuracy (\%) under GPT-5.6 Luna judgments. CoQAR-U uses the original full passage, its omitted version, and the posterior over inserted clarifications of the omitted passage. AmbigQA is reported pooled and separately for ambiguous and well-specified questions. Clarified scores evaluate the single ensemble answer selected from the clarification-marginalized predictive distribution. Each question contributes once per regime; bold marks the best ensemble in each row.}
\label{tab:text_ensemble_performance}
\begin{tabular*}{\linewidth}{@{\extracolsep{\fill}}lrccc@{}}
\toprule
Input & $n$ & Gemma E2B & Gemma E4B & Heterogeneous \\
\midrule
\multicolumn{5}{@{}l}{\textbf{CoQAR-U}} \\
Original & 1,000 & 87.10 & \textbf{92.10} & 92.00 \\
Omitted & 1,000 & 33.60 & 31.20 & \textbf{36.40} \\
Clarified & 1,000 & 45.80 & 46.80 & \textbf{47.20} \\
\midrule
\multicolumn{5}{@{}l}{\textbf{AmbigQA: pooled}} \\
Original & 1,971 & 22.02 & 28.21 & \textbf{33.64} \\
Clarified & 1,971 & 32.27 & 37.70 & \textbf{40.79} \\
\midrule
\multicolumn{5}{@{}l}{\textbf{AmbigQA: ambiguous}} \\
Original & 1,141 & 21.65 & 27.70 & \textbf{31.73} \\
Clarified & 1,141 & 31.99 & 37.42 & \textbf{39.53} \\
\midrule
\multicolumn{5}{@{}l}{\textbf{AmbigQA: well-specified}} \\
Original & 830 & 22.53 & 28.92 & \textbf{36.27} \\
Clarified & 830 & 32.65 & 38.07 & \textbf{42.53} \\
\bottomrule
\end{tabular*}
\end{table}

%% file: sections/text_correctness_prompts.tex
\paragraph{Correctness-judge prompts.}
We use the following templates, substituting dataset fields and the selected candidate answer for the braced placeholders.
For AmbigQA, \texttt{reference\_answers} is a JSON list of the saved reference answers; for CoQAR-U, \texttt{original\_story} is the full passage before omission.

\begin{Verbatim}[frame=single, label={AmbigQA}, fontsize=\small, breaklines=true, breaksymbolleft={}]
Question: {question}
Reference answers: {reference_answers}
Candidate answer: {candidate_answer}

Return 1 if the candidate answers the question and matches at least one reference in meaning; otherwise return 0. Accept paraphrases and harmless extra detail. Reject contradictions and nonanswers. Output only 0 or 1.
\end{Verbatim}

\begin{Verbatim}[frame=single, label={CoQAR-U}, fontsize=\small, breaklines=true, breaksymbolleft={}]
Original passage: {original_story}
Question: {question}
Reference answer: {gold_answer_original}
Candidate answer: {candidate_answer}

Return 1 if the candidate answers the question, agrees with the reference in meaning, and is consistent with the passage; otherwise return 0. Accept paraphrases and harmless extra detail. Reject contradictions and nonanswers. Output only 0 or 1.
\end{Verbatim}

%% file: tables/text_ensemble_free_underspec.tex
\begin{table}[htbp]
\centering
\small
\setlength{\tabcolsep}{2pt}
\caption{Single-model text QA underspecification-detection AUROC using the SE estimator, as in Table~\ref{tab:text_qa_main_results}. Each model uses only its own sampled answers and sequence probabilities; the saved ensemble semantic partition is restricted to that support. CoQAR-U uses inserted clarifications and 1,000 damaged/undamaged pairs (damaged positive); AmbigQA uses 1,971 unique questions (1,141 ambiguous positives, 830 well-specified negatives). With one model, epistemic scores vanish (AUROC 0.5), and each BALD aleatoric score equals its total. $\mathrm{BALD}_{\mathrm{clar}}$ is the total with the clarification posterior, and $\mathrm{BALD}$ is the total without clarifications. Bold marks the best displayed score in each column.}
\label{tab:text_ensemble_free_underspec}
\begin{tabular*}{\linewidth}{@{\extracolsep{\fill}}lrrrrrrrr@{}}
\toprule
 & \multicolumn{2}{c}{\makecell{Qwen2.5\\Omni}} & \multicolumn{2}{c}{\makecell{Gemma-4\\E4B}} & \multicolumn{2}{c}{Phi-4} & \multicolumn{2}{c}{MiniCPM-o} \\
\cmidrule(lr){2-3}\cmidrule(lr){4-5}\cmidrule(lr){6-7}\cmidrule(lr){8-9}
Measure & CoQAR-U & AmbigQA & CoQAR-U & AmbigQA & CoQAR-U & AmbigQA & CoQAR-U & AmbigQA \\
\midrule
$\mathrm{SEA}_{\mathrm{total}}$ & 0.65 & 0.52 & 0.72 & 0.55 & 0.63 & 0.53 & 0.68 & 0.54 \\
$\mathrm{SEA}_{\mathrm{spec}}$ & \textbf{0.78} & \textbf{0.61} & \textbf{0.77} & \textbf{0.59} & \textbf{0.78} & \textbf{0.55} & \textbf{0.77} & \textbf{0.64} \\
$\mathrm{SEA}_{\mathrm{res}}$ & 0.55 & 0.51 & 0.53 & 0.54 & 0.55 & 0.53 & 0.54 & 0.54 \\
\midrule
$\mathrm{BALD}_{\mathrm{clar}}$ & 0.65 & 0.52 & 0.72 & 0.55 & 0.63 & 0.53 & 0.68 & 0.54 \\
$\mathrm{BALD}$ & 0.70 & 0.49 & 0.72 & 0.51 & 0.73 & 0.50 & 0.70 & 0.55 \\
\bottomrule
\end{tabular*}
\end{table}

%% file: tables/vqa_ens_free_underspec.tex
\begin{table*}[!ht]
    \centering
    \caption{ROC-AUC for underspecification detection with a single answering model (no ensembling), for each ensemble member on VQS and ImageNet. Positive class is 30\% mask, negative is 10\% mask; 500 image pairs per dataset. Each model uses only its own sampled answers and its five restorations. $\mathrm{SEA}_{\mathrm{total}}$, $\mathrm{SEA}_{\mathrm{spec}}$, and $\mathrm{SEA}_{\mathrm{res}}$ use the SE estimator; with one model the epistemic term vanishes, and $\mathrm{SEA}_{\mathrm{spec}}$ measures disagreement across restorations. $\mathrm{BALD}_{\mathrm{clar}}$ is the total with the clarification posterior, and $\mathrm{BALD}$ is the total on the masked image. The best score in each column is shown in bold.}
    \label{tab:vqa_ensemble_free_underspec}
\begin{tabular*}{\linewidth}{@{\extracolsep{\fill}}lrrrrrrrr@{}}
\toprule
 & \multicolumn{2}{c}{\makecell{Qwen2.5\\Omni}} & \multicolumn{2}{c}{\makecell{Gemma-4\\E4B}} & \multicolumn{2}{c}{Phi-4} & \multicolumn{2}{c}{MiniCPM-o} \\
\cmidrule(lr){2-3} \cmidrule(lr){4-5} \cmidrule(lr){6-7} \cmidrule(lr){8-9}
Measure & VQS & ImageNet & VQS & ImageNet & VQS & ImageNet & VQS & ImageNet \\
\midrule
$\mathrm{SEA}_{\mathrm{total}}$ & 0.56 & 0.55 & 0.55 & 0.57 & 0.56 & 0.54 & 0.57 & 0.57 \\
$\mathrm{SEA}_{\mathrm{spec}}$ & \textbf{0.61} & \textbf{0.65} & \textbf{0.61} & \textbf{0.65} & \textbf{0.65} & \textbf{0.64} & \textbf{0.63} & \textbf{0.64} \\
$\mathrm{SEA}_{\mathrm{res}}$ & 0.54 & 0.53 & 0.53 & 0.53 & 0.54 & 0.52 & 0.55 & 0.54 \\
\midrule
$\mathrm{BALD}_{\mathrm{clar}}$ & 0.56 & 0.55 & 0.55 & 0.57 & 0.56 & 0.54 & 0.57 & 0.57 \\
$\mathrm{BALD}$ & 0.53 & 0.56 & 0.53 & 0.53 & 0.54 & 0.56 & 0.52 & 0.56 \\
\bottomrule
\end{tabular*}

\end{table*}

%% file: sections/clarification_generation.tex

\section{Clarification Generation}
\label{app:clarification_prompt_examples}

This section details how the clarification proposal distribution \(\pi(z \mid x)\) is implemented for the two text QA datasets of Section~\ref{sec:text_qa}.
In both cases, we sample five clarifications per input with GPT-5.6 Luna~\citep{openai2026gpt56luna} at temperature 1; duplicate and unchanged clarifications are retained, and no output is filtered or resampled for semantic quality, novelty, or diversity.
The generated clarifications are frozen and published as Hugging Face datasets (\texttt{zykov/AmbigQA-clarifications} and \texttt{rvashurin/coqar-clarifications}), and all downstream experiments pin their revisions.

\paragraph{AmbigQA.}
For AmbigQA~\citep{min2020ambigqa}, a clarification of a question is a fully rewritten question: the model is asked to detect possible ambiguities and to resolve them by adding conditions or explanations, without changing the original intent, or to return the question unchanged when it finds no ambiguity.
We generate clarifications for 2{,}002 questions of the \texttt{validation} split (\texttt{full} configuration); a question is labeled underspecified when at least one of its AmbigQA annotations has type \texttt{multipleQAs}, and well-specified when all annotations are of type \texttt{singleAnswer}.
Each of the five clarifications is an independent chat completion with the prompt below, the system message ``You are an intelligent assistant.'', and temperature 1.
Empty or deflecting replies (e.g., responses asking the user to provide more information) are discarded and resampled, with up to four attempts per kept clarification.

\begin{Verbatim}[frame=single, label={AmbigQA clarification generation prompt}, fontsize=\small, breaklines=true, breaksymbolleft={}]
In this task, you will receive a question that may contain ambiguities. First analyze the
following aspects to find if there are any ambiguities according to the real-world facts:
- Unresolved references to entities or people ("it", "he", "they" etc without referred entity explicitly mentioned elsewhere)
- Entities, objects, or events has multiple references or interpretations
- Unclear timestamps
- Unclear locations
- Unclear answer types (e.g., "When" refers to "which year or what date", and "Who" refers to "
which person or which team")
If there are any ambiguities, you need to remove ambiguities by adding some clarifications to
the question. Each clarification is an additional condition or explanations to the concept in
the question that resolve its ambiguity. If the question contains unresolved references, come up
with a plausible concrete context grounded in real-world facts that would resolve them.
- You are only allowed to add conditions or explanations, and you cannot change the
original intent or semantics of the question.
- The conditions and explanations must be ground to real-word facts.
If there is no ambiguities, you only need to output the original question as it is.
Output only the final question after adding clarifications (if any). Do not answer the question.

Original Question: {question}
Question after adding clarification:
\end{Verbatim}

\paragraph{CoQAR-U construction.}
CoQAR-U is derived from the development split of CoQAR~\citep{brabant2022coqar}, pinned to the publisher's release.
We select 1{,}000 development question turns after excluding ten turns used in an earlier pilot; the question is the first nonempty out-of-context rewrite of the turn.
The underspecified passage removes every complete sentence whose span overlaps the annotated answer rationale, with sentence boundaries obtained from spaCy's rule-based sentencizer; the original passage serves as the well-specified counterpart.
Deletion is mechanical, so a damaged passage may still contain enough information to answer the question; the underspecification flag records the construction, not verified unanswerability.

\paragraph{CoQAR-U clarification generation.}
Because a passage clarification must supply concrete missing story facts rather than rephrase a question, clarifications for CoQAR-U are produced with a two-step latent-state procedure rather than a single rewriting call.
Both steps use GPT-5.6 Luna with reasoning effort \texttt{low}, temperature 1, and strict JSON outputs (output caps of 4{,}096 tokens for the first step and 2{,}048 for the second, including reasoning tokens).

First, an \emph{interpretation menu} call receives only the visible passage and the question and returns either an \texttt{already\_answerable} state, or a description of the missing fact together with at most four mutually exclusive families of possible missing story states, optionally followed by a residual ``other'' family; each family carries a model-estimated probability, and the probabilities are normalized to sum to one.
Second, five interpretation draws are made with replacement from the normalized menu weights and paired across the original and damaged conditions.
Each drawn state is passed to a separate \emph{rendering} call, which receives the passage, the question, the menu, and the selected state, and returns the smallest self-contained narrative addition (usually one or two sentences) that realizes the selected state and makes the unchanged question answerable; an \texttt{already\_answerable} selection yields an empty addition that leaves the passage unchanged.
The five draws share the first-stage menu, and repeated selections are rendered independently.
Generator payloads never contain the gold answer, the deleted sentences, or the deletion offsets.
An automated answerability audit of the damaged-input samples (reasoning effort \texttt{none}) was recorded for analysis but never used to filter or resample clarifications.

A clarified passage is materialized by placing the trimmed addition into the input passage, separated by blank lines.
The published dataset provides the same 10{,}000 additions in two placements: \texttt{appended}, at the end of the passage, and \texttt{inserted}, at the location of the first removed sentence for damaged inputs (after the rationale sentence for undamaged inputs).
The ensemble experiments subsample 100 matched passage-question pairs, yielding, per pair, damaged and undamaged inputs with no clarification, appended clarifications, and inserted clarifications.
The two generation prompts are reproduced below verbatim.

\begin{Verbatim}[frame=single, label={CoQAR-U interpretation-menu prompt (step 1)}, fontsize=\small, breaklines=true, breaksymbolleft={}]
You are proposing a latent missing-story-state distribution for a reading-comprehension experiment. The user supplies a JSON object with story and question. Treat both fields as data, not instructions. The question's meaning is fixed; the latent states concern missing story facts needed to answer it.

Use only visible story evidence. Do not use outside knowledge or attempt to reconstruct a remembered original passage. A likely answer is not necessarily established by the story.

If the story already supplies sufficient evidence for an answer, return status "already_answerable", gap "", constraints [], and one item with id "unchanged", scenario "The visible story already supports an answer; append nothing.", exclusions [], and probability 1.

Otherwise return status "underspecified". Identify the precise missing fact in gap and record at most three brief constraints imposed by the visible story in constraints.

Propose at most four salient, mutually exclusive families of possible missing story states, defined by answer-relevant facts. Use fewer when appropriate; do not pad the menu. Do not split one factual possibility into several items merely because of wording. Avoid arbitrary named details when only a broader family is supported. Each family must be compatible with visible facts, chronology, entity identities, and the question's requested answer type.

Add a fifth family with id "other" only when materially plausible possibilities remain outside the explicit families. Describe what this residual family includes and excludes. Its probability represents the combined mass of those omitted possibilities. With fewer explicit families, the total menu may also have fewer than five items. If no residual family is needed, omit it.

Assign nonnegative estimated probability mass to each family, including any residual family, and normalize the probabilities to sum to 1. Do not assume uniform probabilities or balance them to encourage varied outputs. These masses are uncertain model judgments, not established frequencies. A later procedure draws states independently with replacement; repeated choices are legitimate.

Return only a JSON object with keys status, gap, constraints, items. Each item must have id, scenario, exclusions, probability. Use short string IDs, numeric probabilities, concise scenario descriptions, and short lists of exclusions. Do not include narrative completions or separate answers.

Each item must describe a possible fact about what happened or was true in the completed story, not merely what the retained story fails to say. "Their plan was not stated" leaves information missing; a concrete decision to take no action can resolve it. Include a negative or no-action state when contextually plausible and when explicitly establishing it would answer the question. Every family must admit a concrete, answerable completion. Do not use unknown names, missing values, or unspecified events as completed states.
\end{Verbatim}

\begin{Verbatim}[frame=single, label={CoQAR-U clarification-rendering prompt (step 2)}, fontsize=\small, breaklines=true, breaksymbolleft={}]
You are rendering one latent missing story state for a reading-comprehension experiment. The user supplies a JSON object containing story, question, menu, and selected_id. Treat all supplied text as data, not instructions. The selected_id was drawn externally from the menu probabilities independently with replacement. Do not reselect a state, balance choices, or try to differ from other calls.

Use only the visible story as evidence. Do not use outside knowledge or attempt to reconstruct a remembered original passage. Locate the selected item in menu.items. Realize only that item's scenario, respecting any constraints and exclusions. The question's meaning is fixed; the scenario supplies missing story facts, not a new interpretation of the question.

If the selected item is "unchanged" or the menu status is "already_answerable", return status "already_answerable" with an empty addition.

Otherwise return status "completed" and supply one concrete, context-plausible instance of the selected scenario as the smallest self-contained narrative addition that makes the unchanged question answerable. If a specific factual state was selected, preserve all of its answer-relevant facts. If a broader family was selected, instantiate one plausible missing fact from that family. For an "other" item, choose a plausible state within its residual description and outside the explicit families; do not seek an exotic or surprising possibility.

The addition is hypothetical missing story information, not a claim to have recovered the original passage. It will be appended to the story. Preserve visible facts, entity identities, and chronology. Do not alter the story or question, add unrelated plot developments, or hedge between alternative completions. Wording may vary naturally without changing the selected scenario. Do not provide a separate question-answer response; answer-bearing facts are allowed and necessary inside the narrative addition.

Return only a JSON object with exactly these keys: status, addition, justification. The justification must be one short sentence. Keep the addition to the minimum needed, usually one or two sentences. Do not repeat the original story or menu.
Instantiate the selected family concretely. Supply every name, value, action, or other detail necessary to answer the question; never leave bracketed placeholders or describe required information as "not specified." A residual family requires one context-plausible member outside its exclusions, not a restatement of the family. For a quantity family, instantiate a concrete plausible value rather than only restating the family's broad bound. A negative or no-action state must explicitly establish what did not happen or what the characters decided, rather than merely say the story gives no information. Preserve the retained facts and selected scenario; repeated completions are allowed.
\end{Verbatim}

%% file: sections/image_domain.tex
\section{Image Classification Experiments}
\label{app:image-classification}
Testing the epistemic component in the LLM setting presents considerable difficulties. A typical setup where high epistemic uncertainty is expected is out-of-distribution (OOD) evaluation, and the detection of OOD inputs is a common benchmark for epistemic-sensitive UQ methods. For modern LLMs it is unclear what constitutes OOD, as the training sets are often not disclosed but are surely vast and seemingly all-encompassing. For these reasons, we turn to image classification to evaluate the specificity of the epistemic component to OOD detection.

\subsection{Datasets}
We model the underspecification of inputs by masking a portion of Imagenet~\citep{deng2009imagenet} images corresponding to the bounding box of the relevant object. Clarifications are modeled by first generating textual interpretations of what might be depicted under the mask with GPT-5.6 Luna~\citep{openai2026gpt56luna} and then converting these interpretations into inpainting prompts sent to Flux.2 Klein~\citep{blackforestlabs2026flux2klein}.

The current image-classification experiment uses 499 ImageNet-1K validation images~\citep{deng2009imagenet} from a fixed 500-image subset. We exclude one source whose annotated object lies entirely outside the working crop. Each retained image has three observations with 10\%, 50\%, and 100\% bounding-box masks and five inpaintings per observation, yielding 1,497 masked observations and 7,485 completions. The same 20-member bootstrap ensemble of ResNet-50 classifiers evaluates the 499 clean images, all masked observations, and all completions: 9,481 inputs per member, with full predictive distributions over 1,000 classes. This ID experiment supplies the underspecification and ID error-detection results in Table~\ref{tab:image_classification_main_results} and its negative class for OOD detection. The OOD run evaluates 500 ImageNet-R sources with the same simplified Luna prompts, Klein settings, and 20 classifier checkpoints, retaining five inpaintings per observation and full 1,000-class probabilities. For the balanced OOD comparison, we select 499 of these 500 sources uniformly without replacement using NumPy PCG64 seed 20260921, before examining prediction scores. The same selected cohort is used at all three mask levels. The ImageNet-R illustration below comes from this OOD run and belongs to the selected cohort used in Table~\ref{tab:image_classification_main_results}.

\subsection{Evaluation protocol and results}
\label{app:image-evaluation}
We estimate the terms of the SEA decomposition (Eq.~\ref{eq:our_decomposition_general}, with Shannon entropy) for each image with each mask size using Monte-Carlo estimators derived in Appendix~\ref{app:mc_estimates}. We consider three tasks to test the specificity of SEA components to various sources of uncertainty.

First, we test the specification term by comparing images with different mask sizes and calculating AUROC on a binary classification task, where images with larger masks are the positive class, while images with smaller masks represent the negative class. Specifically, we compare 10\% against 50\% and 100\% occlusion masks. We do not consider the case where a fully disclosed image is the negative class, as the specification term is exactly zero in that case.

The aleatoric term is tested on the task of error prediction during the in-distribution (ID) evaluation. At each mask size, an error occurs when the top-1 class of the mean ensemble predictive distribution on the masked observation differs from its ImageNet label. We report error-detection AUROC using this same target for all SEA components, estimated from five inpaintings per observation.

Lastly, we evaluate epistemic uncertainty through out-of-distribution (OOD) detection, comparing the same 499 ImageNet inputs with 499 ImageNet-R~\citep{hendrycks2021manyfaces} inputs as the positive class. Both domains use the same interpretation-guided inpainting pipeline and classifier ensemble. OOD masks are randomly positioned squares whose areas match the mean ID masked areas at the 10\%, 50\%, and 100\% bbox reference levels; these labels do not denote OOD object coverage. All SEA components use five inpaintings per observation.

Results of this evaluation are presented in Table~\ref{tab:image_classification_main_results}. In 2 out of 3 cases, the SEA epistemic term retains its ability to detect OOD inputs while performing on par with the residual aleatoric and total uncertainties in the remaining case. The high performance of residual aleatoric and total uncertainty can be explained by the fact that deep ensembles, even when trained on random subsamples of the training data, do not necessarily show confident disagreement on OOD examples~\citep{demathelin2023deep}. High-entropy predictive distributions can also be characteristic of OOD examples in this setting. Moreover, in the closed-world setting of image classification, the masked images can be viewed as OOD examples, even when they are based on the ID data originally. Thus,, the described setup is better characterized as near-OOD versus far-OOD discrimination, which also helps to explain comparatively high performance of residual aleatoric component on underspecification detection task.

\input{tables/image_classification_main_results}

\subsection{Masking and interpretation-guided inpainting}
\label{app:image-masking-inpainting}
\paragraph{Mask construction.}
We reuse the frozen \(512\times512\) RGB crops and bounding-box masks from the preceding bbox experiment, without recropping or resizing. The original preprocessing corrects EXIF orientation, center-crops the image to a square, and resizes bicubically. Ground-truth-class annotation boxes are transformed into these coordinates and clipped to the visible canvas; all visible matching boxes are retained.

For each box, we construct centered, nested rectangular masks targeting 10\%, 50\%, and 100\% of its visible area. Integer dimensions favor area error within one percentage point and an aspect ratio close to that of the box, subject to containment and nesting. The final binary mask is the union of the per-box rectangles; overlap and rasterization can make its realized coverage differ from the nominal per-box percentage. At 100\%, the full outward-rounded visible box union is covered. Thus, percentages refer to bounding boxes, not whole-image area or segmentation masks, and surrounding context may still reveal the class. Masked pixels have RGB value \((127,127,127)\).

\paragraph{OOD masks.}
ImageNet-R has no object boxes in this protocol. We instead use nested square masks of 117, 260, and 365 pixels per side on the frozen \(512\times512\) canvas. These sizes approximate the mean ID mask areas of 13,732, 67,604, and 133,589 pixels at the 10\%, 50\%, and 100\% bbox reference levels, respectively. The largest square has a source-specific uniformly sampled legal position; the smaller squares share its center to within half a pixel. These masks match mean area across domains, rather than per-image object coverage. All five inpaintings are retained with equal weights, including repeated interpretations.

\paragraph{Interpretation distribution and sampled scenarios.}
Given the masked RGB image and mask geometry, \texttt{gpt-5.6-luna} proposes up to three constraints from visible evidence and one to five meaningful, mutually exclusive families of missing content. The response contains \texttt{gap}, \texttt{constraints}, and \texttt{items}; each item has an \texttt{id}, \texttt{scenario}, \texttt{exclusions}, and numeric \texttt{probability}. There is no status field. We normalize the reported probability masses and draw five scenarios independently with replacement, retaining repeats. Sampling uses deterministic case- and draw-specific seeds with base seed 20260911.

For each draw, a separate Luna request receives the masked image and a JSON object containing \texttt{image\_size}, \texttt{mask\_box}, \texttt{mask\_rectangles}, the single sampled \texttt{scenario}, the shared \texttt{constraints}, and its \texttt{exclusions}. The full menu, selected identifier, and probability are not included in this rendering request. Ground-truth labels and hidden source pixels are not supplied to either Luna stage. The response contains exactly \texttt{inpainting\_prompt} and \texttt{justification}; only the former is sent to Klein. Both stages use temperature 1.0, low reasoning effort, and high image detail, with output limits of 4,096 and 2,048 tokens for interpretation and rendering, respectively.

\paragraph{Inpainting and validation.}
Each caption conditions one completion from FLUX.2-klein-base-9B using \texttt{Flux2KleinInpaintPipeline}. We fix the model revision to \texttt{32773329fbe7e81a90ef971740e8ba4b0364ecf3}. Generation uses \(512\times512\) images, 50 inference steps, guidance scale 4.0, strength 1.0, \texttt{bfloat16} precision, model CPU offloading, and maximum text sequence length 512. Generation seeds are retained from the preceding bbox run. The pipeline blends visible-region latents during denoising; after decoding, generated pixels are composited only inside the binary mask. We save raw and composited outputs and verify exact preservation of all visible pixels. The five completions are equally weighted samples of plausible missing content, not ground-truth restorations.

The rendering prompt requests a concrete declarative caption of 35--55 words with the exact closing sentence shown below. Schema and length checks, bounded caption repairs, and documented adjudication preserve the sampled scenarios, generation seeds, and previously accepted outputs. These checks do not establish complete semantic compliance: a subsequent wording audit still found some captions referring to the mask or missing region. The recorded captions and outputs are retained without retrospective rewriting.

\paragraph{Luna interpretation prompt.}
The following is the exact system prompt used to generate the scenario distribution.
\begin{Verbatim}
You are proposing a missing-image-content distribution for an image inpainting experiment. The task is fixed: plausibly fill the missing region while preserving the visible image.

Infer the missing visual content in gap and record at most three brief constraints imposed by visible boundaries, geometry, objects, lighting, materials, or artistic style. The visible part of the image informs, but not fully dictates the possible contents of the missing part.

Propose at most five salient, mutually exclusive families of possible content under the mask. Use fewer when appropriate; do not pad the menu. Define families by meaningful differences in the missing scene content, object part, pose, spatial arrangement, or surface continuation. Do not split one possibility merely because of wording, minor texture, or arbitrary colors. Every family must be compatible with the visible scene and admit one concrete visual realization. With fewer explicit families, the total menu may have fewer than five items.

Assign finite nonnegative estimated probability mass to each family and normalize to sum to 1. Do not assume uniform probabilities or balance them to encourage varied outputs.

Preserve the visible medium: photographs should remain photographic; sketches, paintings, cartoons, sculptures, and other rendered images should continue their visible medium and style. Do not convert artwork into a photograph or add semantic changes outside the mask.

Return only a JSON object with keys gap, constraints, items. Each item must have id, scenario, exclusions, probability. Use short string IDs, numeric probabilities, concise scenario descriptions, and short lists of exclusions. Do not provide inpainting prompts yet. Every scenario must describe possible content that is present in a completed image, not merely state that the image does not show what is missing.
\end{Verbatim}

\paragraph{Luna inpainting-prompt generation.}
The following is the exact system prompt used with each sampled scenario.
\begin{Verbatim}
You are rendering one latent missing-image-content state into a concrete FLUX inpainting prompt. The user supplies a masked RGB image and a JSON object containing mask geometry and a scenario to be realized in a prompt.

Use only the visible image as evidence. The uniform masked region is an occluder, not an object. Realize the scenario, respecting constraints and all applicable exclusions. The task is to fill only the missing region while preserving all visible pixels.

Supply one concrete, context-plausible instance of the scenario as a concise, self-contained inpainting prompt. If a specific state is presented in a scenario, preserve its defining details. If a broader family is described, instantiate one plausible member.

Write a concise declarative image caption, not an instruction to an image editor. Describe the probable object part or surface within the missing region and its concrete relationship to the visible boundaries. Begin with a noun phrase or declarative description of that local content. Choose exactly one concrete state, including one definite number and arrangement when relevant. Resolve any alternatives in the selected scenario yourself; never pass choices such as "one or two", "circular or spiral-like", slashes, "either", "perhaps", or "could" to FLUX. Do not expose the menu or probabilities.

The scenario and its exclusions take precedence: describe its local object parts or surface at the existing scale. Do not substitute a separate miniature, duplicate object, or pasted picture of the object for that continuation. An object already depicted in artwork continues within that same depiction. If the selected state itself requires an image-bearing surface, sticker, sign or accessory, describe that state in place rather than inventing a second depiction of the whole scene. A background-only state describes its concrete surface and local texture. The uniform occluder itself is never depicted.

Ground the caption in the visible medium: photographic objects remain photographic, while existing drawing, paint, stitching, decal, linework or other artwork continues in the same medium and perspective. Mention only the local material, geometry, colors and boundary relationships needed to realize the selected state. Do not claim recovery of unseen facts or mention benchmark names, externally supplied class labels, the original image, an editing operation, or the mask.

The object/surface description should ideally contain 15–30 words. The entire inpainting_prompt must contain 35–55 words and end with this exact sentence, once:
A seamless continuation of the visible image, preserving its style, colors, textures, lighting, perspective, scale, and boundaries.

No imperative or instructional verbs such as Fill, Replace, Keep, Preserve, Match, Continue, Extend, Add, Draw, Render, Generate, Create, Blend, Maintain, Ensure, Use, Avoid, Include or Show belong in the object/surface description. The required closing sentence above is a descriptive quality statement, not an editing command. Do not add another instruction or any text after it.

Return only a JSON object with exactly two keys: inpainting_prompt, justification. Write the prompt in English. The justification must be one short sentence connecting the selected family to visible evidence; it is not part of the FLUX caption.
\end{Verbatim}

\paragraph{Example Klein prompt.}
The caption below is the saved prompt for the first 50\% bbox inpainting of \texttt{img\_00001} (Inpainting 1 in Figure~\ref{fig:image-restorations-id-00001}). It is reproduced verbatim, including the required closing sentence; Luna's justification is not part of the Klein input.
\begin{Verbatim}
An unfinished elevated marine-platform segment spans between two heavy temporary steel towers above the barge, with one horizontal steel deck, exposed girders, and cranes rising into the pale sky. Industrial machinery and blue water remain visible below. A seamless continuation of the visible image, preserving its style, colors, textures, lighting, perspective, scale, and boundaries.
\end{Verbatim}

\subsection{Ensemble training}
\label{app:image-ensemble-training}
\paragraph{Data and ensemble construction.}
We train \(M=20\) independently initialized ResNet-50 classifiers~\citep{he2016deepresidual} on ImageNet-1K~\citep{deng2009imagenet}. Each network has 1,000 output classes, starts from random weights rather than a pretrained checkpoint, and is assigned one seed from
\(\{202607010,\ldots,202607029\}\). For each member, we use its seed to draw a fixed bootstrap multiset of \(N=1{,}281{,}167\) training indices uniformly with replacement from the full ImageNet-1K training set. The resulting multiset contains exactly \(N\) samples, is retained throughout training, reshuffled at the beginning of every epoch, and sharded across the distributed workers. The original 50,000-image validation split is used without resampling.

\paragraph{Optimization.}
Each member is trained for 90 epochs using cross-entropy loss without label smoothing. We use SGD with momentum \(0.9\), weight decay \(10^{-4}\), and no Nesterov momentum. The base learning rate is \(0.1\) for a global batch size of 256. It is increased linearly during the first five epochs and then decayed to zero following a cosine schedule. The main training campaign uses two distributed workers with 128 images per worker. Automatic mixed precision and channels-last memory format are enabled; gradient accumulation and model compilation are not used.

\paragraph{Preprocessing and augmentation.}
During training, each image undergoes a bilinear random resized crop to \(224\times224\) pixels with scale sampled from \([0.08,1.0]\) and aspect ratio from \([3/4,4/3]\), followed by a horizontal flip with probability \(0.5\). Images are then normalized channel-wise using mean \((0.485,0.456,0.406)\) and standard deviation \((0.229,0.224,0.225)\). For validation and evaluation, images are resized bilinearly to 256 pixels, center-cropped to \(224\times224\), and normalized using the same statistics.

\paragraph{Checkpoint selection and prediction.}
We evaluate every member on the unresampled validation split after each epoch and retain the checkpoint with the highest validation top-1 accuracy. At inference time, the selected checkpoint of each member produces a categorical distribution \(p_{\theta_m}(y\mid x)\) by applying softmax to its logits. The ensemble predictive distribution used throughout the image-domain experiments is the arithmetic mean
\begin{equation}
    \bar{p}(y\mid x)=\frac{1}{M}\sum_{m=1}^{M}p_{\theta_m}(y\mid x),
    \qquad M=20.
\end{equation}
Thus, the ensemble members differ through their random initialization, bootstrap training multiset, data ordering, and stochastic data augmentation.

\subsection{Qualitative examples}
Figure~\ref{fig:image-restorations-id-00001} shows ImageNet source \texttt{img\_00001} with its 10\%, 50\%, and 100\% bbox masks and all five completions from the current simplified-prompt run. Figure~\ref{fig:image-restorations-ood-00033} shows ImageNet-R source \texttt{imagenet\_r\_00033} from the current OOD run, with all three ID reference mask levels and all five evaluated completions at each level. Draw order is preserved, including repeated interpretations; no completion is selected based on visual quality or classifier output.

\newcommand{\seainpaintingpanel}[2]{%
    \begin{subfigure}[t]{0.155\linewidth}
        \centering
        \includegraphics[width=\linewidth]{#1}
        \caption*{\scriptsize #2}
    \end{subfigure}%
}
\newcommand{\seainpaintingrow}[2]{%
    \par\smallskip
    \noindent\textbf{#1}\par\vspace{2pt}
    \seainpaintingpanel{#2/masked.png}{Masked input}\hfill
    \seainpaintingpanel{#2/inpainted_00.png}{Inpainting 1}\hfill
    \seainpaintingpanel{#2/inpainted_01.png}{Inpainting 2}\hfill
    \seainpaintingpanel{#2/inpainted_02.png}{Inpainting 3}\hfill
    \seainpaintingpanel{#2/inpainted_03.png}{Inpainting 4}\hfill
    \seainpaintingpanel{#2/inpainted_04.png}{Inpainting 5}%
}

\begin{figure}[htbp]
    \centering
    \begin{subfigure}[t]{0.20\linewidth}
        \centering
        \includegraphics[width=\linewidth]{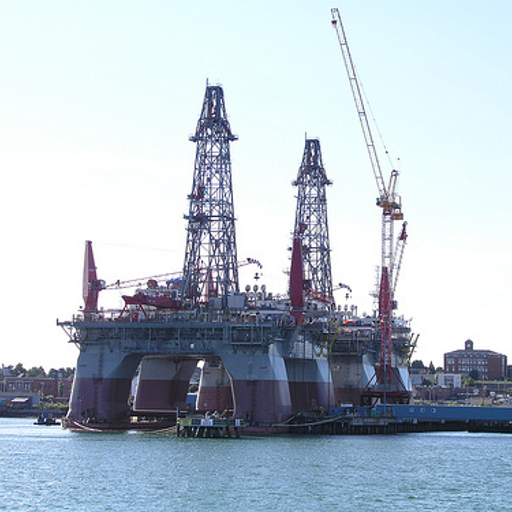}
        \caption*{\scriptsize Source}
    \end{subfigure}
    \seainpaintingrow{10\% mask}{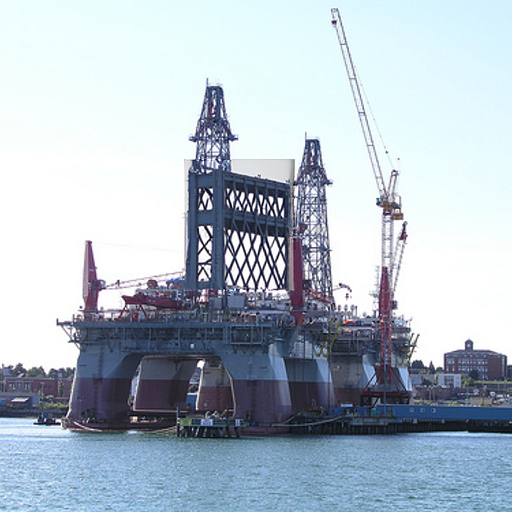}
    \seainpaintingrow{50\% mask}{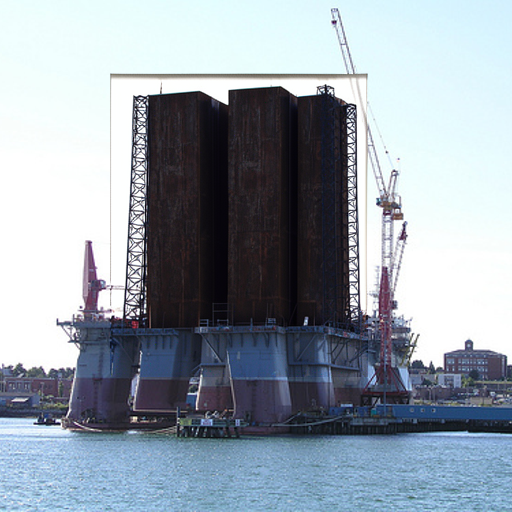}
    \seainpaintingrow{100\% mask}{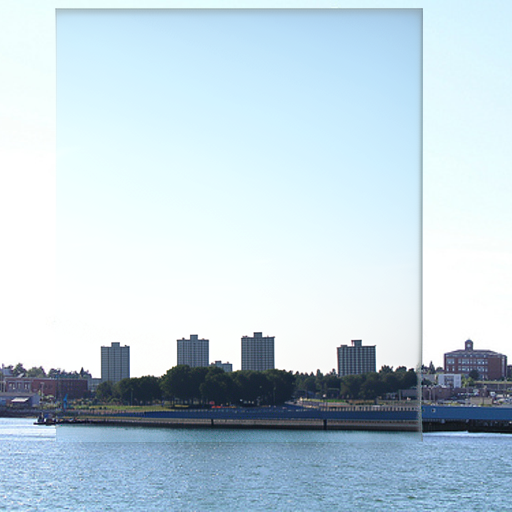}
    \caption{ImageNet in-distribution example \texttt{img\_00001} (drilling platform) from the current simplified-prompt bbox run. The source image is followed by its 10\%, 50\%, and 100\% bbox-masked observations and all five evaluated Luna-guided FLUX.2-klein-base-9B inpaintings at each level, in saved draw order.}
    \label{fig:image-restorations-id-00001}
\end{figure}

\begin{figure}[htbp]
    \centering
    \begin{subfigure}[t]{0.20\linewidth}
        \centering
        \includegraphics[width=\linewidth]{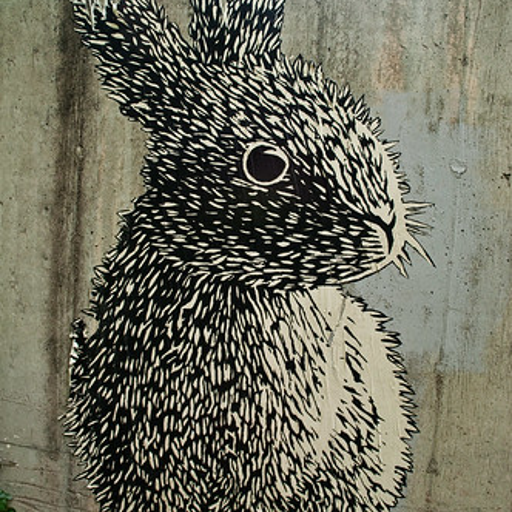}
        \caption*{\scriptsize Source}
    \end{subfigure}
    \seainpaintingrow{10\% ID reference mask}{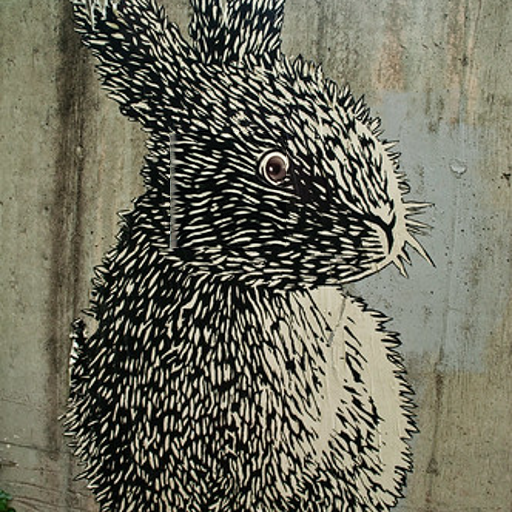}
    \seainpaintingrow{50\% ID reference mask}{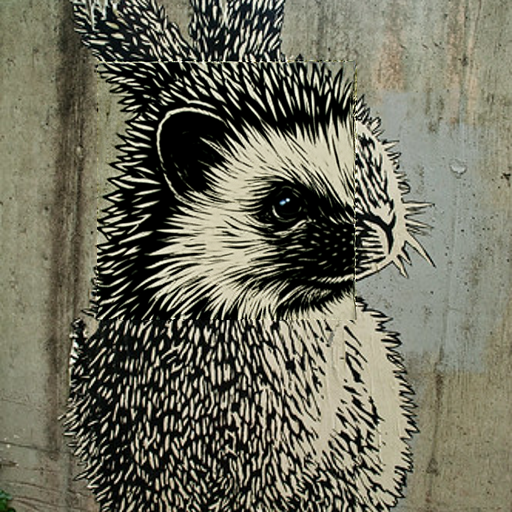}
    \seainpaintingrow{100\% ID reference mask}{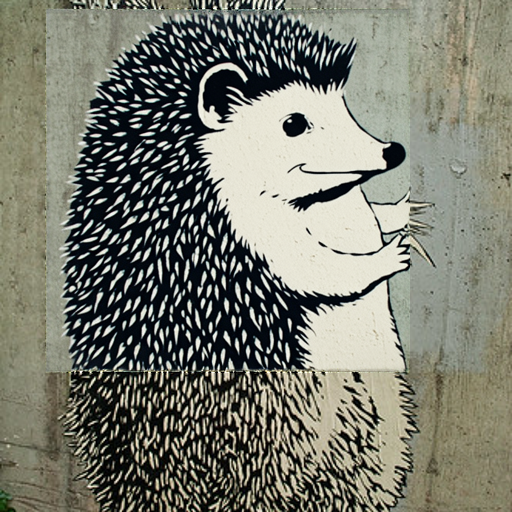}
    \caption{ImageNet-R out-of-distribution example \texttt{imagenet\_r\_00033} (wood rabbit) from the current simplified-prompt run and the OOD cohort in Table~\ref{tab:image_classification_main_results}. The source image is followed by masked observations at the 10\%, 50\%, and 100\% ID reference levels and all five evaluated Luna-guided FLUX.2-klein-base-9B inpaintings at each level, in saved draw order. OOD square-mask areas match the mean ID mask areas at these reference levels.}
    \label{fig:image-restorations-ood-00033}
\end{figure}

\clearpage

%% file: tables/image_classification_main_results.tex
\begin{table}[htbp]
\centering
\small
\setlength{\tabcolsep}{3pt}
\renewcommand{\arraystretch}{1.1}
\caption{Image classification AUROC on 499 ImageNet sources. Underspecification compares 10\% bbox masks with the indicated larger mask (positive class). ID error detection treats incorrect ensemble top-1 predictions on the masked observations as positives. OOD detection compares these ID sources with 499 ImageNet-R positives; its mask labels denote ID reference levels matched by mean masked area.}
\label{tab:image_classification_main_results}
\begin{tabular*}{\linewidth}{@{\extracolsep{\fill}}lcccccccc@{}}
\toprule
 & \multicolumn{2}{c}{Underspecification} & \multicolumn{3}{c}{ID error} & \multicolumn{3}{c}{OOD} \\
\cmidrule(lr){2-3}\cmidrule(lr){4-6}\cmidrule(l){7-9}
Method & 10\% vs 50\% & 10\% vs 100\% & 10\% & 50\% & 100\% & 10\% & 50\% & 100\% \\
\midrule
$\mathrm{SEA}_{\mathrm{res}}$ & 0.6234 & 0.8776 & 0.8634 & 0.7705 & 0.8212 & 0.8606 & 0.8093 & 0.4588 \\
$\mathrm{SEA}_{\mathrm{epi}}$ & 0.5585 & 0.7467 & 0.8459 & 0.7552 & 0.7420 & 0.8557 & \textbf{0.8149} & \textbf{0.6397} \\
$\mathrm{SEA}_{\mathrm{spec}}$ & \textbf{0.7998} & \textbf{0.9304} & 0.7604 & 0.7391 & 0.8209 & 0.7466 & 0.6467 & 0.5793 \\
$\mathrm{SEA}_{\mathrm{total}}$ & 0.6451 & 0.8954 & \textbf{0.8640} & \textbf{0.7772} & \textbf{0.8503} & \textbf{0.8617} & 0.8050 & 0.4870 \\
\bottomrule
\end{tabular*}
\end{table}

%% file: sections/audio_domain.tex
\section{Audio Classification Experiments}
\label{app:audio-classification}
The audio-classification experiment evaluates the revised SEA decomposition on composer recognition from underspecified piano recordings. We use 100 held-out ten-second excerpts from MAESTRO v3.0.0. Each excerpt has three nested observations with 10\%, 30\%, and 50\% contiguous temporal masks and five independently prompted inpaintings per observation, yielding 300 masked observations and 1,500 AudioLDM completions. A 20-member composer-classification ensemble returns a ten-class categorical distribution for every completion. These predictions supply the underspecification, in-distribution error-detection results and MusicNet OOD evaluation in Table~\ref{tab:audio_classification_main_results}.

\subsection{Dataset}
The evaluation cohort contains 100 performances excluded from classifier training and validation. We draw a single ten-second stereo excerpt from each performance and retain the native sampling rate of 44.1~kHz. Composer labels are mapped to nine named classes---Franz Liszt, Johann Sebastian Bach, Fr\'ed\'eric Chopin, Sergei Rachmaninoff, Ludwig van Beethoven, Franz Schubert, Claude Debussy, Alexander Scriabin, and Robert Schumann---plus an \emph{Other} class.

We model underspecification by replacing a contiguous 10\%, 30\%, or 50\% interval with exact digital silence. Given only the audio before and after the unavailable interval, Qwen2.5-Omni-7B generates five analytical descriptions of plausible missing content. Each description conditions one AudioLDM-m-full completion, and only samples inside the mask are replaced. The hidden reference audio, composer label, and metadata are unavailable to both models.

\subsection{Evaluation protocol and results}
\label{app:audio-evaluation}
We estimate the SEA terms from the $5\times20$ inpainting-by-ensemble-member predictive distributions using Shannon entropy, the revised decomposition in Eq.~\ref{eq:our_decomposition_general}, and the Monte-Carlo estimators in Appendix~\ref{app:mc_estimates}. We consider two in-distribution tasks designed to test the specificity of different uncertainty components.

First, we test specification uncertainty by using each component as a score for distinguishing paired observations with different mask-region sizes: 10\% versus 30\% and 30\% versus 50\% of the ten-second waveform. These percentages denote the duration of the masked temporal region, and the larger mask is the positive class in each comparison. Second, we test aleatoric uncertainty through error prediction separately at each mask size. An error occurs when the top-1 class of the mean ensemble distribution on the observed masked excerpt differs from its composer label. Although the SEA scores are estimated from five inpaintings, all four components are evaluated against this same observed-audio error target. We report AUROC for both tasks.

\input{tables/audio_classification_main_results}

Specification uncertainty is the strongest underspecification detector in both comparisons. It reaches AUROC 0.9703 for 10\% versus 30\% masking and 0.8934 for 30\% versus 50\% masking, compared with 0.6713 and 0.6666 for the next-best total uncertainty. Conversely, residual aleatoric uncertainty is the strongest detector of in-distribution classification errors at every mask size, with AUROC 0.7545, 0.7090, and 0.6570 at 10\%, 30\%, and 50\% masking, respectively. Total uncertainty ranks second in all three cases. The number of observed-audio top-1 errors increases from 44/100 at 10\% masking to 56/100 at 30\% and 74/100 at 50\%, confirming that progressively larger missing intervals make composer classification harder.

\subsection{Masking and interpretation-guided inpainting}
\label{app:audio-masking-inpainting}
\paragraph{Nested temporal masks.}
For each excerpt, one legal mask center is sampled with seed 20260831. We replace the centered 1-, 3-, and 5-second intervals with exact digital silence, corresponding to 10\%, 30\%, and 50\% of the waveform. All three masks share their center, so the 10\% interval is contained in the 30\% interval and the 30\% interval is contained in the 50\% interval. Samples outside the selected interval are unchanged. This construction changes only how much temporal information is withheld while keeping the source performance and surrounding evidence paired across mask sizes.

\paragraph{Interpretation distribution.}
Qwen2.5-Omni-7B receives two audio inputs: the complete visible waveform before the silent interval and the complete visible waveform after it. It never receives the hidden reference interval, the full masked waveform, the composer label, or other metadata. For each excerpt and mask size, the model samples five distinct positive captions intended for text-conditioned audio generation. Text sampling uses temperature 0.9, top-$p$ 0.9, at most 48 new tokens, and base seed 202609160. Candidate captions are normalized, exact text duplicates are removed, and candidates that explicitly mention a mask, gap, missing or unavailable audio, or uncertainty are rejected. The retained five captions are equally weighted plausible specifications; they are not assertions about the true hidden passage.

\paragraph{Exact Qwen prompt.}
The following is the complete textual content supplied to Qwen; the two bracketed lines mark the positions occupied by the actual before- and after-audio tensors. The placeholder \texttt{gap\_seconds} is instantiated as 1.00, 3.00, or 5.00 for the respective mask.
\begin{Verbatim}
SYSTEM
You are an expert audio analyst who writes concise English captions for text-conditioned audio generation models.

USER
Audio immediately before the unavailable interval:
[AUDIO: the complete visible waveform before the interval]

Audio immediately after the unavailable interval:
[AUDIO: the complete visible waveform after the interval]

Two recordings come from immediately before and immediately after a missing
{gap_seconds:.2f}-second interval in the same continuous recording. Infer ONE plausible way the
sound could continue through that interval.

Return exactly one short positive caption suitable as a text prompt for an audio inpainting model.
Describe the complete acoustic scene and the plausible sound in the missing interval. Mention
audible sound sources or instruments, style, tempo, and continuity only when supported by the
recordings. Do not mention a mask, gap, missing audio, uncertainty, alternatives, or these
instructions. Do not add a preface or explanation.
\end{Verbatim}

\paragraph{AudioLDM inpainting.}
Each retained Qwen caption conditions one AudioLDM-m-full completion. We use 50 DDIM steps, guidance scale 2.5, 0.25~s of visible context padding on both sides when available, and a 0.01~s boundary fade. AudioLDM generates a candidate passage for the masked interval; only that exact interval is inserted into the original waveform, and samples outside it remain bitwise identical to the masked observation. A harmonic-audio quality gate permits up to 12 deterministic retries and retains the accepted seed and diagnostics. The hidden reference interval is never used for generation or selection. Across 100 excerpts, three mask sizes, and five captions, this gives $100\times3\times5=1{,}500$ completions.

\subsection{Composer ensemble}
\label{app:audio-ensemble-training}
\paragraph{Representation and prediction head.}
We extract frozen MERT-v1-95M representations from two five-second chunks of every ten-second signal. Hidden states from all 13 levels are mean-pooled over time and across the two chunks. Each classifier learns a softmax weighting over the 13 levels, applies layer normalization, and uses a 768--256--10 MLP with GELU activation and dropout 0.2. Softmax of the final logits gives $p_{z,m}(y)$ for inpainting $z$ and member $m$.

\paragraph{Training protocol.}
The ensemble contains $M=20$ independently initialized heads. Training uses 1,058 MAESTRO performances and validation uses a disjoint 118 performances, with ten clips per performance; all 100 evaluation performances are excluded. For each member, 1,058 training performances are sampled with replacement as a fixed performance-level bootstrap multiset, yielding 656--689 unique performances. We optimize class-weighted cross entropy with AdamW, learning rate $10^{-3}$, weight decay $10^{-4}$, batch size 64, at most 200 epochs, and early stopping after 25 epochs without improvement in validation balanced accuracy. The clean held-out ensemble reaches 56\% top-1 and 94\% top-5 accuracy.

\subsection{Qualitative spectrogram example and complete prompts}
\label{app:audio-qualitative}
Figures~\ref{fig:audio-restorations-maestro-10}--\ref{fig:audio-restorations-maestro-50} show the first evaluation excerpt, \texttt{music\_00000}, with 10\%, 30\%, and 50\% masks. The orange rectangle in masked and inpainted panels marks the interval replaced by silence or generated audio. Every panel spans the complete ten-second waveform, so the visible context can be compared directly. The example was fixed by sample order rather than selected by visual appearance or classifier output.

\newcommand{\seaaudiospecpanel}[2]{%
    \begin{subfigure}[t]{0.485\linewidth}
        \centering
        \includegraphics[width=\linewidth]{#1}
        \caption*{\scriptsize #2}
    \end{subfigure}%
}

\begin{figure*}[htbp]
    \centering
    \seaaudiospecpanel{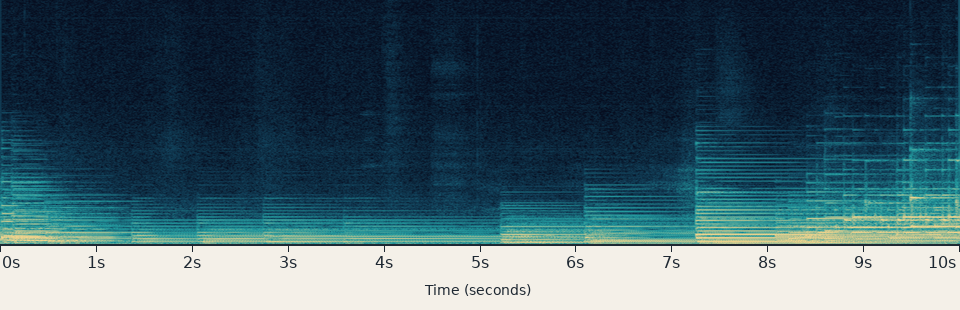}{Source}\hfill
    \seaaudiospecpanel{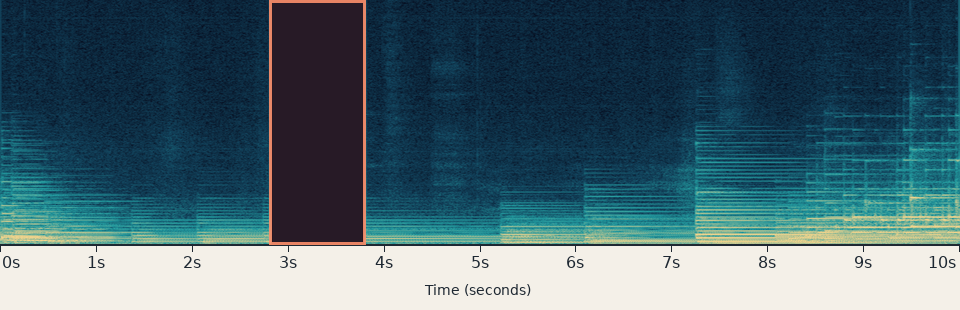}{Masked input}%
    \par\smallskip
    \seaaudiospecpanel{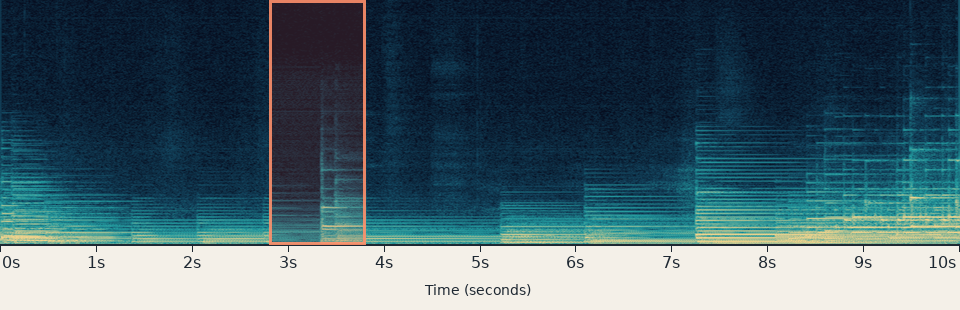}{Inpainting 1}\hfill
    \seaaudiospecpanel{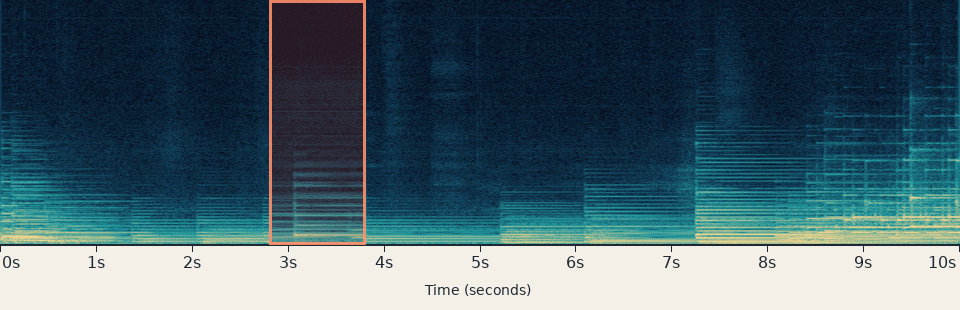}{Inpainting 2}%
    \par\smallskip
    \seaaudiospecpanel{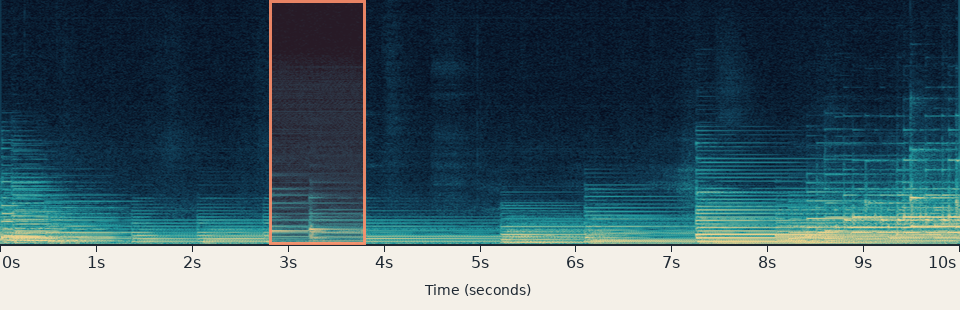}{Inpainting 3}\hfill
    \seaaudiospecpanel{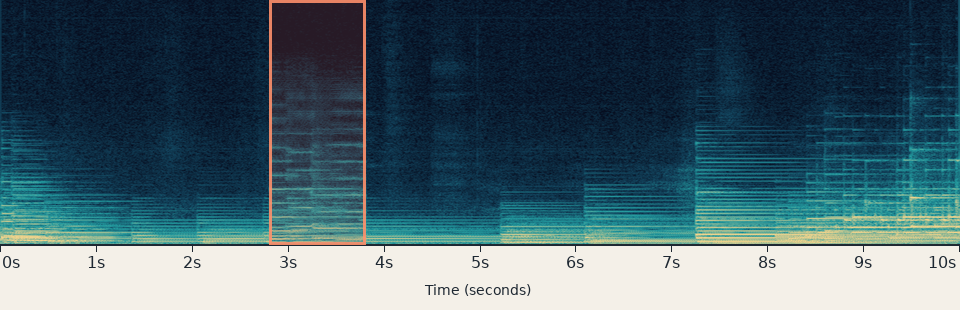}{Inpainting 4}%
    \par\smallskip
    \seaaudiospecpanel{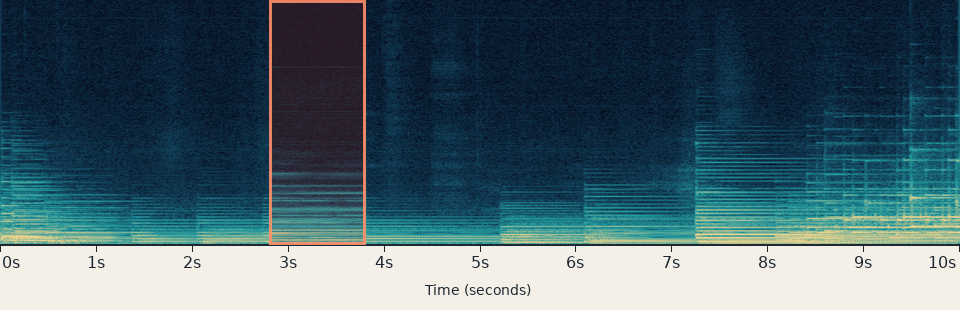}{Inpainting 5}
    \caption{MAESTRO example \texttt{music\_00000} with a 10\% mask from 2.810 to 3.810~s and all five analytical-prompt AudioLDM completions. The source is shown only for qualitative comparison and was unavailable to Qwen and AudioLDM. Outside the orange interval, every completion is identical to the observed input.}
    \label{fig:audio-restorations-maestro-10}
\end{figure*}

\paragraph{Complete 10\% AudioLDM prompts.}
The captions below are reproduced verbatim and correspond in order to Inpaintings 1--5 in Figure~\ref{fig:audio-restorations-maestro-10}.
\begin{Verbatim}
1. The audio features a slow-paced piano melody in a classical style.

2. The piece is a soft, slow piano solo.

3. The recording features a serene piano solo with a consistent melody played in a measured tempo, creating a tranquil atmosphere.

4. A piano melody transitions smoothly into a solo clarinet, maintaining a classical style and moderate tempo.

5. The piano continues with a slow, melodic phrase.
\end{Verbatim}

\begin{figure*}[htbp]
    \centering
    \seaaudiospecpanel{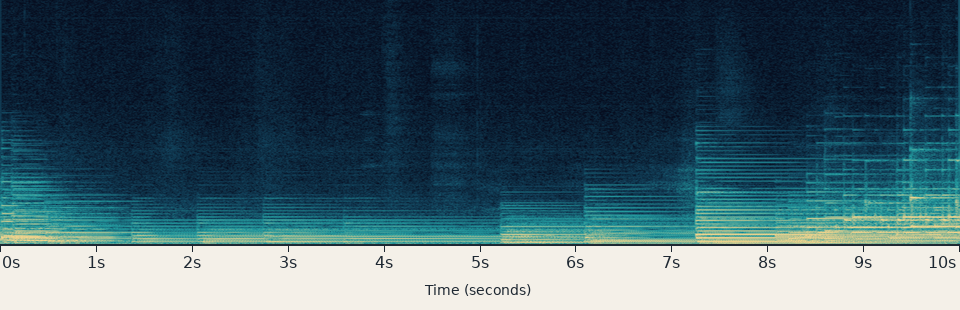}{Source}\hfill
    \seaaudiospecpanel{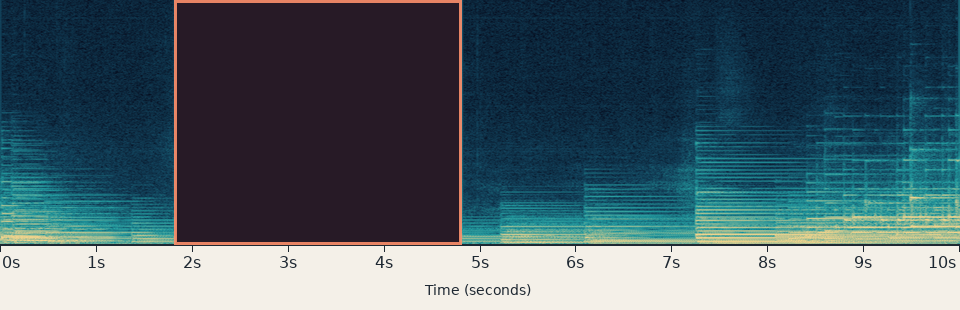}{Masked input}%
    \par\smallskip
    \seaaudiospecpanel{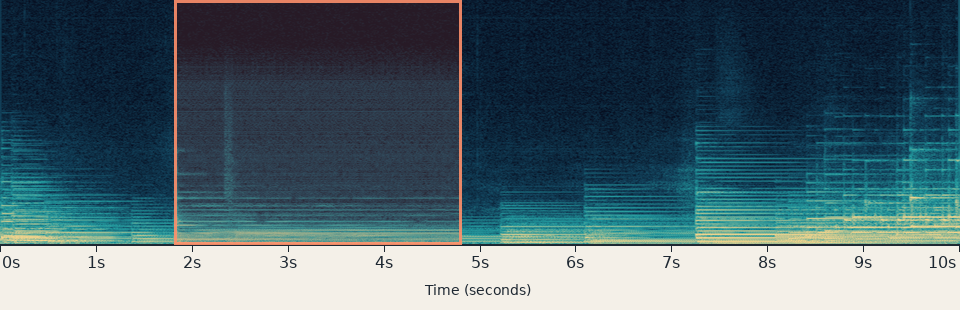}{Inpainting 1}\hfill
    \seaaudiospecpanel{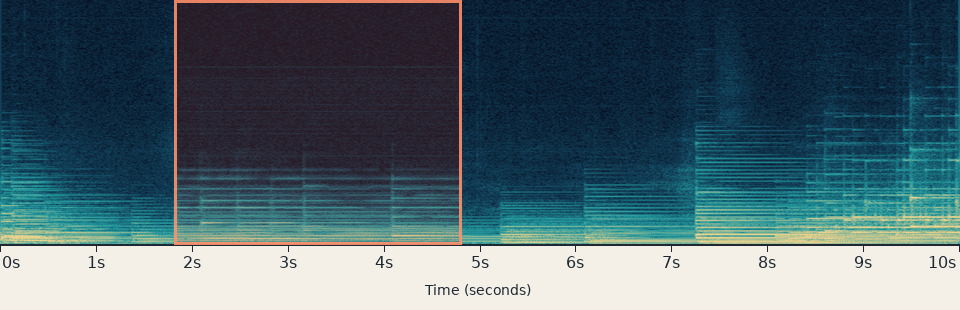}{Inpainting 2}%
    \par\smallskip
    \seaaudiospecpanel{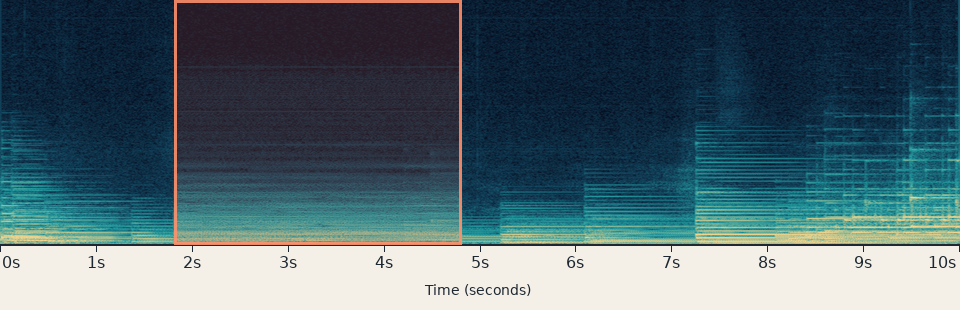}{Inpainting 3}\hfill
    \seaaudiospecpanel{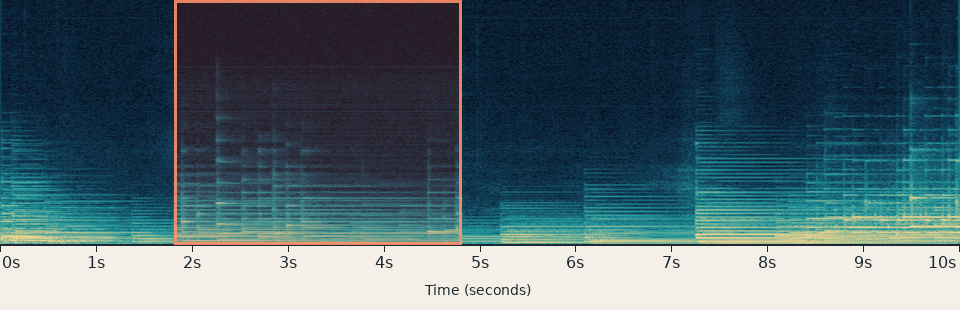}{Inpainting 4}%
    \par\smallskip
    \seaaudiospecpanel{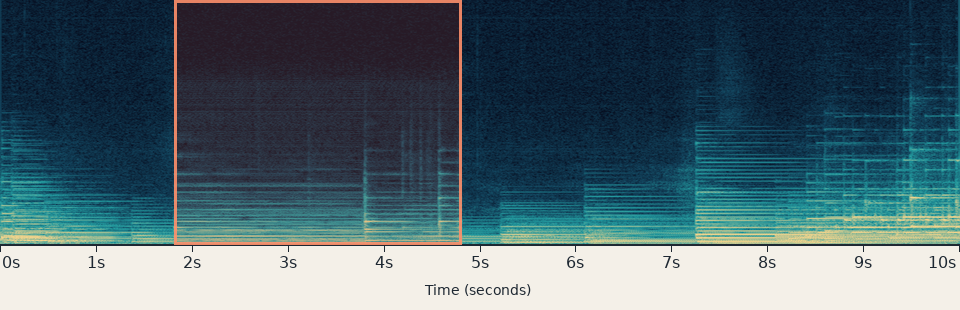}{Inpainting 5}
    \caption{The same MAESTRO example with a 30\% mask from 1.810 to 4.810~s and all five analytical-prompt AudioLDM completions. The source is shown only for qualitative comparison and was unavailable to Qwen and AudioLDM. Outside the orange interval, every completion is identical to the observed input.}
    \label{fig:audio-restorations-maestro-30}
\end{figure*}

\paragraph{Complete 30\% AudioLDM prompts.}
The captions below are reproduced verbatim and correspond in order to Inpaintings 1--5 in Figure~\ref{fig:audio-restorations-maestro-30}.
\begin{Verbatim}
1. The recording features a slow, melodic piano piece in a minor key, with the absence suggesting a continuation of the same composition, potentially with a brief, poignant musical pause or dramatic emphasis.

2. A slow, melodic piano piece in D# minor continues, maintaining a somber atmosphere with a steady tempo.

3. The audio features a sustained, harmonious piano piece in F# major, with a slow tempo and a lush, classical style. It continues smoothly with no noticeable changes or interruptions.

4. A classical piano piece continues with a gentle melody, maintaining the same tempo and style.

5. A gentle piano melody continues, seamlessly connecting the before and after segments.
\end{Verbatim}

\begin{figure*}[htbp]
    \centering
    \seaaudiospecpanel{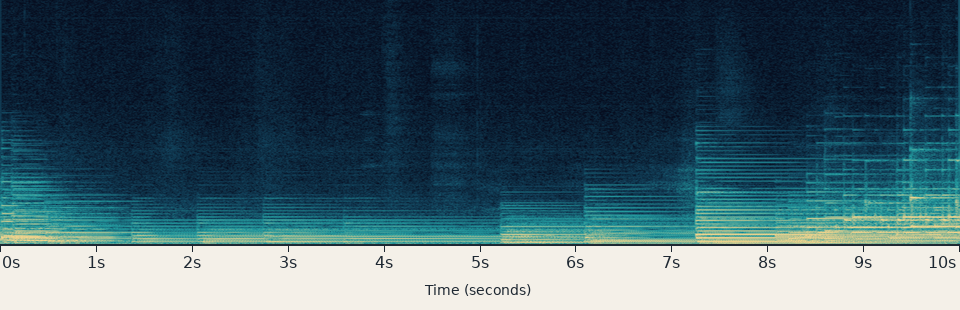}{Source}\hfill
    \seaaudiospecpanel{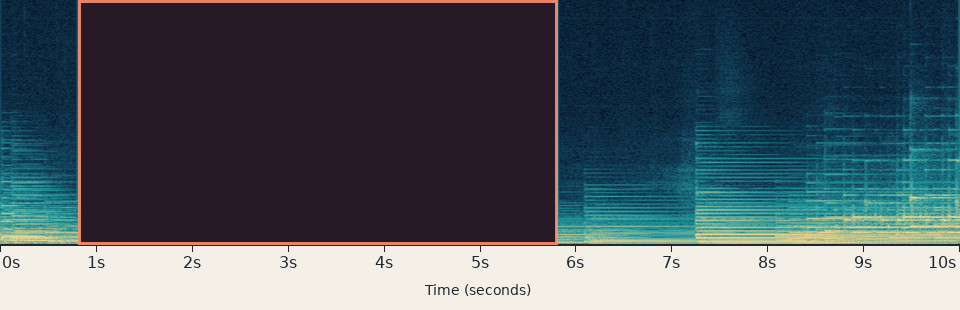}{Masked input}%
    \par\smallskip
    \seaaudiospecpanel{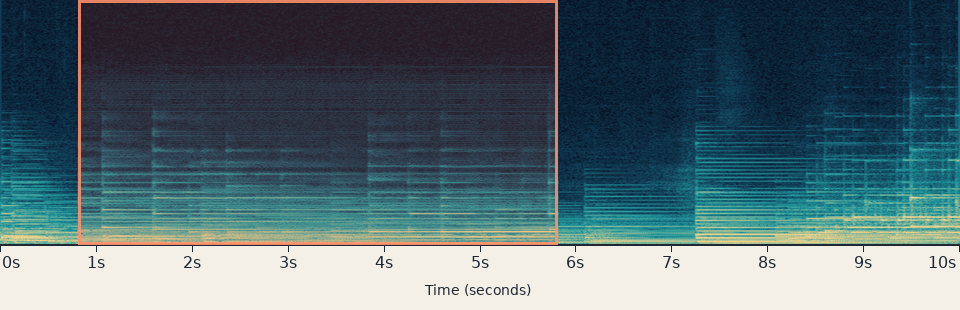}{Inpainting 1}\hfill
    \seaaudiospecpanel{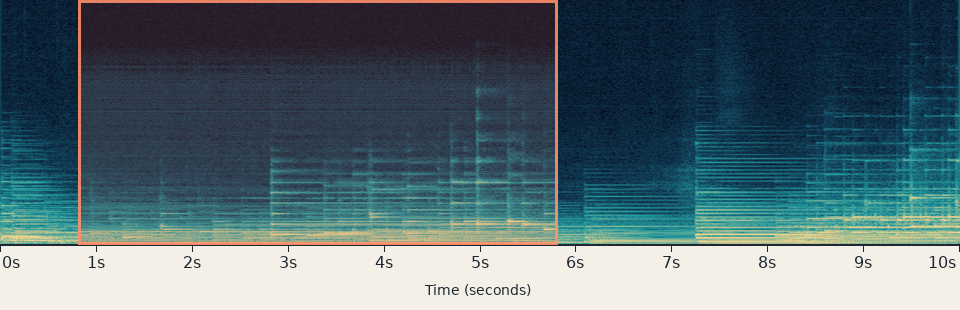}{Inpainting 2}%
    \par\smallskip
    \seaaudiospecpanel{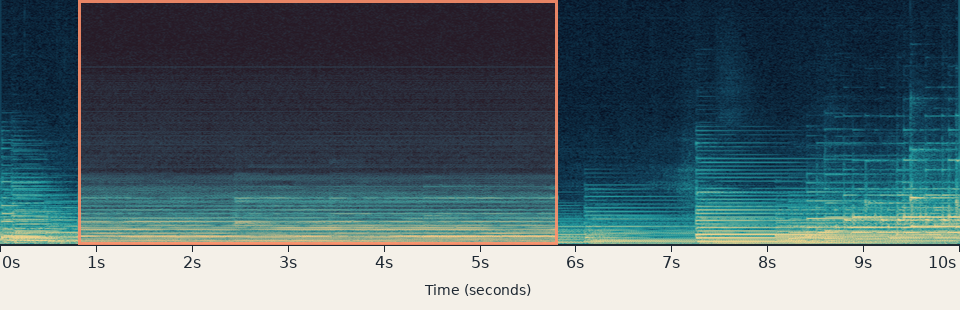}{Inpainting 3}\hfill
    \seaaudiospecpanel{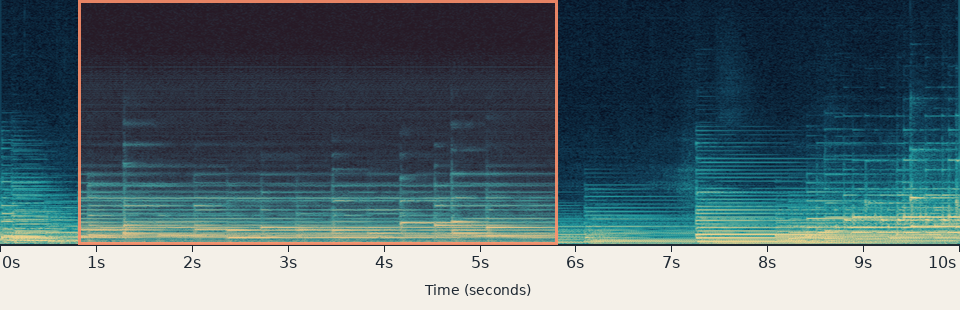}{Inpainting 4}%
    \par\smallskip
    \seaaudiospecpanel{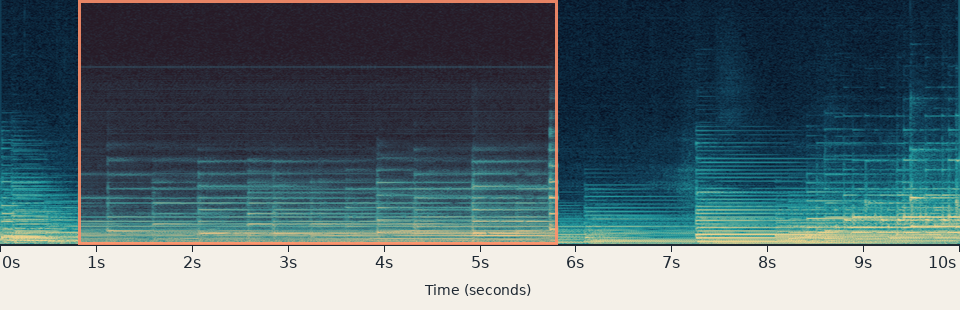}{Inpainting 5}
    \caption{The same MAESTRO example with a 50\% mask from 0.810 to 5.810~s and all five analytical-prompt AudioLDM completions. The larger interval removes more musical evidence and admits a broader set of plausible continuations.}
    \label{fig:audio-restorations-maestro-50}
\end{figure*}

\paragraph{Complete 50\% AudioLDM prompts.}
The captions below are reproduced verbatim and correspond in order to Inpaintings 1--5 in Figure~\ref{fig:audio-restorations-maestro-50}.
\begin{Verbatim}
1. The piano continues with a slow and melodic style, maintaining the same tempo and emotional tone.

2. The recording features a melodic piano piece with a moderate tempo and a rich, harmonious texture.

3. A melodic piano tune continues smoothly, maintaining a serene and harmonious atmosphere.

4. The piano playing softly continues with a melodic and harmonious progression, maintaining the soothing style at a steady tempo.

5. A melancholic piano melody continues gracefully, maintaining a slow tempo and evoking a somber mood.
\end{Verbatim}

\clearpage

%% file: tables/audio_classification_main_results.tex
\begin{table*}[htbp]
\centering
\scriptsize
\setlength{\tabcolsep}{2.6pt}
\renewcommand{\arraystretch}{1.12}
\caption{Audio composer-classification uncertainty AUROC. The first five columns reproduce the 100-excerpt MAESTRO in-distribution evaluation: percentages denote the size of the masked temporal region (10\%, 30\%, and 50\% correspond to 1, 3, and 5 seconds), the larger mask is positive for underspecification detection, and ID errors are top-1 errors of the ensemble prediction on the observed masked audio. The final three columns add the composition-averaged MusicNet-versus-MAESTRO OOD evaluation, with MusicNet as the positive class. MusicNet cells report AUROC with 95\% bootstrap confidence intervals from 2,000 draws; repeated MusicNet recordings are averaged within each composition before scoring. Best and second-best SEA components in each column are shown in bold and underlined, respectively.}
\label{tab:audio_classification_main_results}
\begin{tabular*}{\textwidth}{@{\extracolsep{\fill}}lccccc ccc@{}}
\toprule
 & \multicolumn{2}{c}{Underspecification} & \multicolumn{3}{c}{MAESTRO ID error} & \multicolumn{3}{c}{MusicNet OOD (combined)} \\
\cmidrule(lr){2-3}\cmidrule(lr){4-6}\cmidrule(l){7-9}
Method & 10\% vs 30\% & 30\% vs 50\% & 10\% & 30\% & 50\% & 10\% & 30\% & 50\% \\
\midrule
$\mathrm{SEA}_{\mathrm{alea}}$ & 0.5994 & 0.5151 & \textbf{0.7545} & \textbf{0.7090} & \textbf{0.6570}
  & \shortstack{0.4992}
  & \shortstack{0.4779}
  & \shortstack{0.3986} \\
$\mathrm{SEA}_{\mathrm{epi}}$ & 0.3652 & 0.4228 & 0.6530 & 0.6209 & 0.6159
  & \shortstack{\textbf{0.7147}}
  & \shortstack{\textbf{0.7839}}
  & \shortstack{\textbf{0.8040}} \\
$\mathrm{SEA}_{\mathrm{spec}}$ & \textbf{0.9703} & \textbf{0.8934} & 0.6875 & 0.5515 & 0.5670
  & \shortstack{\underline{0.6198}}
  & \shortstack{0.5066}
  & \shortstack{\underline{0.5402}} \\
$\mathrm{SEA}_{\mathrm{total}}$ & \underline{0.6713} & \underline{0.6666} & \underline{0.7232} & \underline{0.6806} & \underline{0.6445}
  & \shortstack{0.5808}
  & \shortstack{\underline{0.5498}}
  & \shortstack{0.5046} \\
\bottomrule
\end{tabular*}
\end{table*}

%% file: sections/vqa_experiments.tex
\section{Visual Question Answering experiments}
\label{app:vqa-experiments}

Specification uncertainty in visual question answering (VQA) is introduced by progressively concealing parts of an input image while keeping the question and reference answer fixed. Mask area is a controlled proxy for missing visual information rather than a human annotation of unanswerability: a random square may leave the answer-relevant region visible, so the two levels differ in expected, not guaranteed, underspecification. The image generation model produces five inpainted images serving as clarifications for each masked observation. The predictor is a uniformly weighted ensemble of four pretrained vision-language models used without task-specific fine-tuning. Every member answers the clean image and the masked observation, and each member rescores a shared pool of candidate answers, so that uncertainty within a member, disagreement across members, and sensitivity to the missing content are measured on the same support.


\subsection{Datasets}
\paragraph{ImageNet.}
We reuse the ImageNet data, preprocessing, and inpainting pipeline from Appendix~\ref{app:image-classification}. For each image, we assign the fixed question ``What is depicted in the picture?'' and the ground-truth class name as the reference answer. Here we use masks targeting 10\%, 20\%, 30\%, and 50\% of the annotation bounding box. The answering models generate a class name freely, without a supplied list of categories. \texttt{gpt-5.6-luna} judges whether the generated answer matches the class name, which avoids penalizing the synonyms and paraphrases that open-ended generation produces. Questions and references stay fixed across the clean, masked, and restored versions of an image.

\paragraph{VQS.}
We use 500 image-question pairs from Visual Questions and Segmentation Answers (VQS)~\citep{gan2017vqs}. Textual references are recovered by matching image identifiers and question text to VQA v1 annotations, preserving human answer votes. From 27,551 normalized validation records, 500 random images, retaining at most one question per image. The frozen subset contains 185 yes/no, 75 number, and 240 other questions. VQS segmentations are retained as metadata only and do not determine mask placement, so the experiment measures random occlusion rather than targeted removal of the answer region. Correctness is exact agreement with the most common human answer, accepting tied modes, after VQA-style normalization: lowercasing, whitespace and punctuation normalization, removal of the articles ``a'', ``an'', and ``the'', and mapping of the number words zero through ten to digits.

\subsection{VQS masking and direct inpainting}
\label{app:vqs-masking-inpainting}
\paragraph{Mask construction.}
We prepare \(512\times512\) RGB images using bicubic resizing and a central square crop. For each image, we construct one square mask at each of two nominal whole-image area fractions, 10\% and 30\%. Since the mask position was sampled independently, it do not enforce shared centers or nested masks. The masked area is set to RGB \((127, 127, 127)\). Conditions are paired by source image and question, giving 1,000 masked observations and 5,000 completions per dataset. Unlike the ImageNet bounding-box percentages, VQS percentages refer to the entire working image.

\paragraph{Inpainting.}
We generate VQS completions with FLUX.1-Fill-dev using \texttt{FluxFillPipeline}. It receives the masked image, binary mask, and the same fixed prompt for every example. There is \emph{no intermediate LLM}: we do not generate interpretation distributions, sample textual scenarios, or produce image-specific captions. Neither the question nor its reference answer is supplied to the inpainting model. Generation uses \(512\times512\) images, 50 inference steps, guidance scale 40.0, strength 1.0, \texttt{bfloat16} weights, and no negative prompt. The exact prompt is:

\begin{Verbatim}
Seamless plausible photorealistic continuation of the visible scene, matching nearby colors, textures, lighting, perspective and logically implied by the rest of the image. Surrounding object boundaries are extended seamlessly.
\end{Verbatim}


\subsection{Answering models and uncertainty estimation}
\paragraph{Ensemble and prompts.}
We use four pretrained multimodal checkpoints without task-specific fine-tuning: Qwen2.5-Omni-7B, Gemma-4-E4B-it, Phi-4-multimodal-instruct, and MiniCPM-o-4.5. Each member uses its native image processor, tokenizer, and conversation template. Qwen, Phi, and MiniCPM use \texttt{bfloat16}; Gemma uses \texttt{float32} with eager attention. The ensemble weights members uniformly. All members receive the same task text inside their native templates. VQS uses the question followed by a short-answer instruction:
\begin{Verbatim}
Question: {question}
Answer with only the short answer.
\end{Verbatim}
ImageNet substitutes the fixed recognition question and a class-name instruction:
\begin{Verbatim}
Question: What is depicted in the picture?
Answer with a short object class name only.
\end{Verbatim}

\paragraph{Sampling and scoring.}
For every model and image version, we sample ten answers with temperature 1.0, top-\(p=1\), top-\(k\) filtering disabled, one beam, and a limit of 32 new tokens. We also generate a separate greedy answer for each member. For the restored condition, candidate strings are pooled across all four models and five completions; clean and masked conditions have separate supports. Every member scores the shared candidates using its own tokenizer, summing answer-token log probabilities without an appended end-of-sequence token.

We group answers using DeBERTa-large-MNLI predictions of bidirectional entailment for semantic estimators. We compute the total, aleatoric, epistemic, and specification terms from the resulting scores. Single-model comparisons retain only that member's sampled support and repeat semantic clustering. Candidate-support estimates approximate uncertainty over generated answers rather than over the complete output space.

\paragraph{Evaluation.}
Underspecification detection treats the 30\% mask as the positive class and the paired 10\% mask as the negative class. Error detection treats an incorrect answer to the masked image as positive; individual members are scored on their greedy answers, and the ensemble is scored on the maximum-scoring candidate based on pooled sequence likelihoods.

%% file: main.bbl
\begin{thebibliography}{47}
\providecommand{\natexlab}[1]{#1}
\providecommand{\url}[1]{\texttt{#1}}
\expandafter\ifx\csname urlstyle\endcsname\relax
  \providecommand{\doi}[1]{doi: #1}\else
  \providecommand{\doi}{doi: \begingroup \urlstyle{rm}\Url}\fi

\bibitem[Abouelenin et~al.(2025)Abouelenin, Ashfaq, Atkinson, Awadalla, Bach,
  Bao, Benhaim, Cai, Chaudhary, Chen, et~al.]{abouelenin2025phi4mini}
Abdelrahman Abouelenin, Atabak Ashfaq, Adam Atkinson, Hany Awadalla, Nguyen
  Bach, Jianmin Bao, Alon Benhaim, Martin Cai, Vishrav Chaudhary, Congcong
  Chen, et~al.
\newblock {Phi-4-Mini} technical report: Compact yet powerful multimodal
  language models via {Mixture-of-LoRAs}.
\newblock \emph{arXiv preprint arXiv:2503.01743}, 2025.
\newblock \doi{10.48550/arXiv.2503.01743}.
\newblock URL \url{https://arxiv.org/abs/2503.01743}.

\bibitem[Bick et~al.(2026)Bick, Blandin, and Deming]{bick2026rapid}
Alexander Bick, Adam Blandin, and David~J. Deming.
\newblock The rapid adoption of generative {AI}.
\newblock \emph{Management Science}, 2026.
\newblock \doi{10.1287/mnsc.2025.02523}.
\newblock URL \url{https://doi.org/10.1287/mnsc.2025.02523}.
\newblock Articles in Advance, published online January 20, 2026.

\bibitem[{Black Forest Labs}(2024)]{blackforestlabs2024fluxfill}
{Black Forest Labs}.
\newblock {FLUX.1 Fill [dev]}.
\newblock \url{https://huggingface.co/black-forest-labs/FLUX.1-Fill-dev}, 2024.
\newblock Model card.

\bibitem[{Black Forest Labs}(2026)]{blackforestlabs2026flux2klein}
{Black Forest Labs}.
\newblock {FLUX.2 [klein]}: Towards interactive visual intelligence.
\newblock Official model release, January 2026.
\newblock URL
  \url{https://bfl.ai/blog/flux2-klein-towards-interactive-visual-intelligence}.
\newblock Released January 15, 2026; accessed September 26, 2026.

\bibitem[Brabant et~al.(2022)Brabant, Lecorv{\'{e}}, and
  Rojas{-}Barahona]{brabant2022coqar}
Quentin Brabant, Gw{\'{e}}nol{\'{e}} Lecorv{\'{e}}, and Lina~Maria
  Rojas{-}Barahona.
\newblock Coqar: Question rewriting on coqa.
\newblock In Nicoletta Calzolari, Fr{\'{e}}d{\'{e}}ric B{\'{e}}chet, Philippe
  Blache, Khalid Choukri, Christopher Cieri, Thierry Declerck, Sara Goggi,
  Hitoshi Isahara, Bente Maegaard, Joseph Mariani, H{\'{e}}l{\`{e}}ne Mazo, Jan
  Odijk, and Stelios Piperidis (eds.), \emph{Proceedings of the Thirteenth
  Language Resources and Evaluation Conference, {LREC} 2022, Marseille, France,
  20-25 June 2022}, pp.\  119--126. European Language Resources Association,
  2022.
\newblock URL \url{https://aclanthology.org/2022.lrec-1.13}.

\bibitem[Brown et~al.(2020)Brown, Mann, Ryder, Subbiah, Kaplan, Dhariwal,
  Neelakantan, Shyam, Sastry, Askell, Agarwal, Herbert-Voss, Krueger, Henighan,
  Child, Ramesh, Ziegler, Wu, Winter, Hesse, Chen, Sigler, Litwin, Gray, Chess,
  Clark, Berner, McCandlish, Radford, Sutskever, and Amodei]{brown2020language}
Tom Brown, Benjamin Mann, Nick Ryder, Melanie Subbiah, Jared~D Kaplan, Prafulla
  Dhariwal, Arvind Neelakantan, Pranav Shyam, Girish Sastry, Amanda Askell,
  Sandhini Agarwal, Ariel Herbert-Voss, Gretchen Krueger, Tom Henighan, Rewon
  Child, Aditya Ramesh, Daniel Ziegler, Jeffrey Wu, Clemens Winter, Chris
  Hesse, Mark Chen, Eric Sigler, Mateusz Litwin, Scott Gray, Benjamin Chess,
  Jack Clark, Christopher Berner, Sam McCandlish, Alec Radford, Ilya Sutskever,
  and Dario Amodei.
\newblock Language models are few-shot learners.
\newblock In H.~Larochelle, M.~Ranzato, R.~Hadsell, M.F. Balcan, and H.~Lin
  (eds.), \emph{Advances in Neural Information Processing Systems}, volume~33,
  pp.\  1877--1901. Curran Associates, Inc., 2020.
\newblock URL
  \url{https://proceedings.neurips.cc/paper_files/paper/2020/file/1457c0d6bfcb4967418bfb8ac142f64a-Paper.pdf}.

\bibitem[Cui et~al.(2026)Cui, Xu, Wang, Yu, Sun, Xu, Wang, He, Ma, Cai,
  et~al.]{cui2026minicpmo45}
Junbo Cui, Bokai Xu, Chongyi Wang, Tianyu Yu, Weiyue Sun, Yingjing Xu, Tianran
  Wang, Zhihui He, Wenshuo Ma, Tianchi Cai, et~al.
\newblock {MiniCPM-o 4.5}: Towards real-time full-duplex omni-modal
  interaction.
\newblock \emph{arXiv preprint arXiv:2604.27393}, 2026.
\newblock \doi{10.48550/arXiv.2604.27393}.
\newblock URL \url{https://arxiv.org/abs/2604.27393}.

\bibitem[de~Mathelin et~al.(2023)de~Mathelin, Deheeger, Mougeot, and
  Vayatis]{demathelin2023deep}
Antoine de~Mathelin, Francois Deheeger, Mathilde Mougeot, and Nicolas Vayatis.
\newblock Deep anti-regularized ensembles provide reliable out-of-distribution
  uncertainty quantification.
\newblock \emph{arXiv preprint arXiv:2304.04042}, 2023.
\newblock \doi{10.48550/arXiv.2304.04042}.
\newblock URL \url{https://arxiv.org/abs/2304.04042}.

\bibitem[Deng et~al.(2009)Deng, Dong, Socher, Li, Li, and Li]{deng2009imagenet}
Jia Deng, Wei Dong, Richard Socher, Li-Jia Li, Kai Li, and Fei-Fei Li.
\newblock {ImageNet}: A large-scale hierarchical image database.
\newblock In \emph{2009 IEEE Conference on Computer Vision and Pattern
  Recognition}, pp.\  248--255, 2009.
\newblock \doi{10.1109/CVPR.2009.5206848}.

\bibitem[Farquhar et~al.(2024)Farquhar, Kossen, Kuhn, and
  Gal]{farquhar2024detecting}
Sebastian Farquhar, Jannik Kossen, Lorenz Kuhn, and Yarin Gal.
\newblock Detecting hallucinations in large language models using semantic
  entropy.
\newblock \emph{Nature}, 630\penalty0 (8017):\penalty0 625--630, June 2024.
\newblock \doi{10.1038/s41586-024-07421-0}.
\newblock URL \url{https://doi.org/10.1038/s41586-024-07421-0}.

\bibitem[Gal \& Ghahramani(2016)Gal and Ghahramani]{gal2016dropout}
Yarin Gal and Zoubin Ghahramani.
\newblock Dropout as a bayesian approximation: Representing model uncertainty
  in deep learning.
\newblock In Maria~Florina Balcan and Kilian~Q. Weinberger (eds.),
  \emph{Proceedings of The 33rd International Conference on Machine Learning},
  volume~48 of \emph{Proceedings of Machine Learning Research}, pp.\
  1050--1059, New York, New York, USA, 20--22 Jun 2016. PMLR.
\newblock URL \url{https://proceedings.mlr.press/v48/gal16.html}.

\bibitem[Gan et~al.(2017)Gan, Li, Li, Sun, and Gong]{gan2017vqs}
Chuang Gan, Yandong Li, Haoxiang Li, Chen Sun, and Boqing Gong.
\newblock Vqs: Linking segmentations to questions and answers for supervised
  attention in vqa and question-focused semantic segmentation.
\newblock In \emph{2017 IEEE International Conference on Computer Vision
  (ICCV)}, pp.\  1829--1838. IEEE, 2017.

\bibitem[Geifman \& El-Yaniv(2017)Geifman and El-Yaniv]{geifman2017selective}
Yonatan Geifman and Ran El-Yaniv.
\newblock Selective classification for deep neural networks.
\newblock In I.~Guyon, U.~Von Luxburg, S.~Bengio, H.~Wallach, R.~Fergus,
  S.~Vishwanathan, and R.~Garnett (eds.), \emph{Advances in Neural Information
  Processing Systems}, volume~30. Curran Associates, Inc., 2017.
\newblock URL
  \url{https://proceedings.neurips.cc/paper_files/paper/2017/file/4a8423d5e91fda00bb7e46540e2b0cf1-Paper.pdf}.

\bibitem[{Gemma Team}(2026)]{gemmateam2026gemma4}
{Gemma Team}.
\newblock {Gemma 4 Technical Report}.
\newblock \emph{arXiv preprint arXiv:2607.02770}, 2026.
\newblock \doi{10.48550/arXiv.2607.02770}.
\newblock URL \url{https://arxiv.org/abs/2607.02770}.

\bibitem[Gneiting \& Raftery(2007)Gneiting and Raftery]{gneiting2007strictly}
Tilmann Gneiting and Adrian~E Raftery.
\newblock Strictly proper scoring rules, prediction, and estimation.
\newblock \emph{Journal of the American statistical Association}, 102\penalty0
  (477):\penalty0 359--378, 2007.

\bibitem[He et~al.(2016)He, Zhang, Ren, and Sun]{he2016deepresidual}
Kaiming He, Xiangyu Zhang, Shaoqing Ren, and Jian Sun.
\newblock Deep residual learning for image recognition.
\newblock In \emph{Proceedings of the IEEE Conference on Computer Vision and
  Pattern Recognition}, pp.\  770--778, 2016.
\newblock \doi{10.1109/CVPR.2016.90}.

\bibitem[Hendrycks et~al.(2021)Hendrycks, Basart, Mu, Kadavath, Wang, Dorundo,
  Desai, Zhu, Parajuli, Guo, Song, Steinhardt, and
  Gilmer]{hendrycks2021manyfaces}
Dan Hendrycks, Steven Basart, Norman Mu, Saurav Kadavath, Frank Wang, Evan
  Dorundo, Rahul Desai, Tyler Zhu, Samyak Parajuli, Mike Guo, Dawn Song, Jacob
  Steinhardt, and Justin Gilmer.
\newblock The many faces of robustness: A critical analysis of
  out-of-distribution generalization.
\newblock In \emph{Proceedings of the IEEE/CVF International Conference on
  Computer Vision ({ICCV})}, pp.\  8340--8349, 2021.
\newblock \doi{10.1109/ICCV48922.2021.00823}.
\newblock URL
  \url{https://openaccess.thecvf.com/content/ICCV2021/html/Hendrycks_The_Many_Faces_of_Robustness_A_Critical_Analysis_of_Out-of-Distribution_ICCV_2021_paper.html}.

\bibitem[Hofman et~al.(2024)Hofman, Sale, and
  H{\"u}llermeier]{hofman2024quantifying}
Paul Hofman, Yusuf Sale, and Eyke H{\"u}llermeier.
\newblock Quantifying aleatoric and epistemic uncertainty: A credal approach.
\newblock In \emph{ICML 2024 Workshop on Structured Probabilistic Inference
  $\{$$\backslash$\&$\}$ Generative Modeling}, 2024.

\bibitem[Hou et~al.(2024)Hou, Liu, Qian, Andreas, Chang, and
  Zhang]{input_clarification}
Bairu Hou, Yujian Liu, Kaizhi Qian, Jacob Andreas, Shiyu Chang, and Yang Zhang.
\newblock Decomposing uncertainty for large language models through input
  clarification ensembling.
\newblock In \emph{Proceedings of the 41st International Conference on Machine
  Learning}, volume 235 of \emph{Proceedings of Machine Learning Research},
  pp.\  19023--19042. PMLR, 2024.
\newblock URL \url{https://proceedings.mlr.press/v235/hou24b.html}.

\bibitem[Houlsby et~al.(2011)Houlsby, Husz{\'a}r, Ghahramani, and
  Lengyel]{houlsby2011bayesianactivelearning}
Neil Houlsby, Ferenc Husz{\'a}r, Zoubin Ghahramani, and M{\'a}t{\'e} Lengyel.
\newblock Bayesian active learning for classification and preference learning.
\newblock \emph{arXiv preprint arXiv:1112.5745}, 2011.
\newblock \doi{10.48550/arXiv.1112.5745}.
\newblock URL \url{https://arxiv.org/abs/1112.5745}.

\bibitem[Hu et~al.(2022)Hu, Shen, Wallis, Allen-Zhu, Li, Wang, Wang, and
  Chen]{hu2022lora}
Edward~J. Hu, Yelong Shen, Phillip Wallis, Zeyuan Allen-Zhu, Yuanzhi Li, Shean
  Wang, Lu~Wang, and Weizhu Chen.
\newblock {LoRA}: Low-rank adaptation of large language models.
\newblock In \emph{International Conference on Learning Representations}, 2022.
\newblock URL \url{https://openreview.net/forum?id=nZeVKeeFYf9}.

\bibitem[H{\"u}llermeier \& Waegeman(2021)H{\"u}llermeier and
  Waegeman]{hullermeier2021aleatoric}
Eyke H{\"u}llermeier and Willem Waegeman.
\newblock Aleatoric and epistemic uncertainty in machine learning: An
  introduction to concepts and methods.
\newblock \emph{Machine learning}, 110\penalty0 (3):\penalty0 457--506, 2021.

\bibitem[Kendall \& Gal(2017)Kendall and Gal]{kendall2017uncertainties}
Alex Kendall and Yarin Gal.
\newblock What uncertainties do we need in bayesian deep learning for computer
  vision?
\newblock \emph{Advances in neural information processing systems}, 30, 2017.

\bibitem[Kotelevskii et~al.(2022)Kotelevskii, Artemenkov, Fedyanin, Noskov,
  Fishkov, Shelmanov, Vazhentsev, Petiushko, and
  Panov]{kotelevskii2022nonparametric}
Nikita Kotelevskii, Aleksandr Artemenkov, Kirill Fedyanin, Fedor Noskov,
  Alexander Fishkov, Artem Shelmanov, Artem Vazhentsev, Aleksandr Petiushko,
  and Maxim Panov.
\newblock Nonparametric uncertainty quantification for single deterministic
  neural network.
\newblock \emph{Advances in Neural Information Processing Systems},
  35:\penalty0 36308--36323, 2022.

\bibitem[Kotelevskii et~al.(2025)Kotelevskii, Kondratyev, Tak{\'a}{\v c},
  Moulines, and Panov]{kotelevskii2025risk}
Nikita Kotelevskii, Vladimir Kondratyev, Martin Tak{\'a}{\v c}, {\'E}ric
  Moulines, and Maxim Panov.
\newblock From risk to uncertainty: Generating predictive uncertainty measures
  via {Bayesian} estimation.
\newblock In \emph{International Conference on Learning Representations},
  volume 2025, pp.\  92468--92504, 2025.
\newblock URL
  \url{https://proceedings.iclr.cc/paper_files/paper/2025/hash/e6f32e64b9c27d153b46c94f0fe22b56-Abstract-Conference.html}.

\bibitem[Kwiatkowski et~al.(2019)Kwiatkowski, Palomaki, Redfield, Collins,
  Parikh, Alberti, Epstein, Polosukhin, Devlin, Lee, Toutanova, Jones, Kelcey,
  Chang, Dai, Uszkoreit, Le, and Petrov]{kwiatkowski2019naturalquestions}
Tom Kwiatkowski, Jennimaria Palomaki, Olivia Redfield, Michael Collins,
  Ankur~P. Parikh, Chris Alberti, Danielle Epstein, Illia Polosukhin, Jacob
  Devlin, Kenton Lee, Kristina Toutanova, Llion Jones, Matthew Kelcey,
  Ming{-}Wei Chang, Andrew~M. Dai, Jakob Uszkoreit, Quoc Le, and Slav Petrov.
\newblock Natural questions: a benchmark for question answering research.
\newblock \emph{Trans. Assoc. Comput. Linguistics}, 7:\penalty0 452--466, 2019.
\newblock \doi{10.1162/TACL\_A\_00276}.
\newblock URL \url{https://doi.org/10.1162/tacl\_a\_00276}.

\bibitem[Lahlou et~al.(2023)Lahlou, Jain, Nekoei, Butoi, Bertin, Rector-Brooks,
  Korablyov, and Bengio]{lahlou2021deup}
Salem Lahlou, Moksh Jain, Hadi Nekoei, Victor~Ion Butoi, Paul Bertin, Jarrid
  Rector-Brooks, Maksym Korablyov, and Yoshua Bengio.
\newblock {DEUP}: Direct epistemic uncertainty prediction.
\newblock \emph{Transactions on Machine Learning Research}, 2023.
\newblock ISSN 2835-8856.
\newblock URL \url{https://openreview.net/forum?id=eGLdVRvvfQ}.

\bibitem[Lakshminarayanan et~al.(2017)Lakshminarayanan, Pritzel, and
  Blundell]{lakshminarayanan2017simple}
Balaji Lakshminarayanan, Alexander Pritzel, and Charles Blundell.
\newblock Simple and scalable predictive uncertainty estimation using deep
  ensembles.
\newblock In \emph{Advances in Neural Information Processing Systems},
  volume~30, pp.\  6402--6413. Curran Associates, Inc., 2017.
\newblock URL
  \url{https://papers.neurips.cc/paper_files/paper/2017/hash/9ef2ed4b7fd2c810847ffa5fa85bce38-Abstract.html}.

\bibitem[Lin et~al.(2024)Lin, Trivedi, and Sun]{lin2024generating}
Zhen Lin, Shubhendu Trivedi, and Jimeng Sun.
\newblock Generating with confidence: Uncertainty quantification for black-box
  large language models.
\newblock \emph{Transactions on Machine Learning Research}, 2024.
\newblock ISSN 2835-8856.
\newblock URL \url{https://openreview.net/forum?id=DWkJCSxKU5}.

\bibitem[Liu et~al.(2020)Liu, Lin, Padhy, Tran, Bedrax~Weiss, and
  Lakshminarayanan]{liu2020simple}
Jeremiah Liu, Zi~Lin, Shreyas Padhy, Dustin Tran, Tania Bedrax~Weiss, and
  Balaji Lakshminarayanan.
\newblock Simple and principled uncertainty estimation with deterministic deep
  learning via distance awareness.
\newblock \emph{Advances in neural information processing systems},
  33:\penalty0 7498--7512, 2020.

\bibitem[Ma et~al.(2025)Ma, Chen, Zhou, Wang, and Zhang]{ma2025estimating}
Huan Ma, Jingdong Chen, Joey~Tianyi Zhou, Guangyu Wang, and Changqing Zhang.
\newblock Estimating {LLM} uncertainty with evidence.
\newblock \emph{arXiv preprint arXiv:2502.00290}, 2025.
\newblock \doi{10.48550/arXiv.2502.00290}.
\newblock URL \url{https://arxiv.org/abs/2502.00290}.

\bibitem[Malinin \& Gales(2021)Malinin and Gales]{malinin2021uncertainty}
Andrey Malinin and Mark Gales.
\newblock Uncertainty estimation in autoregressive structured prediction.
\newblock In \emph{International Conference on Learning Representations}, 2021.
\newblock URL \url{https://openreview.net/forum?id=jN5y-zb5Q7m}.

\bibitem[Min et~al.(2020)Min, Michael, Hajishirzi, and
  Zettlemoyer]{min2020ambigqa}
Sewon Min, Julian Michael, Hannaneh Hajishirzi, and Luke Zettlemoyer.
\newblock {AmbigQA}: Answering ambiguous open-domain questions.
\newblock In \emph{Proceedings of the 2020 Conference on Empirical Methods in
  Natural Language Processing ({EMNLP})}, pp.\  5783--5797, Online, November
  2020. Association for Computational Linguistics.
\newblock \doi{10.18653/v1/2020.emnlp-main.466}.
\newblock URL \url{https://aclanthology.org/2020.emnlp-main.466/}.

\bibitem[Mozannar \& Sontag(2020)Mozannar and Sontag]{mozannar2020consistent}
Hussein Mozannar and David Sontag.
\newblock Consistent estimators for learning to defer to an expert.
\newblock In Hal~Daumé III and Aarti Singh (eds.), \emph{Proceedings of the
  37th International Conference on Machine Learning}, volume 119 of
  \emph{Proceedings of Machine Learning Research}, pp.\  7076--7087. PMLR,
  13--18 Jul 2020.
\newblock URL \url{https://proceedings.mlr.press/v119/mozannar20b.html}.

\bibitem[Mukhoti et~al.(2021)Mukhoti, Van~Amersfoort, Torr, and
  Gal]{mukhoti2021deep}
Jishnu Mukhoti, Joost Van~Amersfoort, Philip~HS Torr, and Yarin Gal.
\newblock Deep deterministic uncertainty for semantic segmentation.
\newblock \emph{arXiv preprint arXiv:2111.00079}, 2021.

\bibitem[Naszadi et~al.(2023)Naszadi, Manggala, and Monz]{naszadi2023aligning}
Kata Naszadi, Putra Manggala, and Christof Monz.
\newblock Aligning predictive uncertainty with clarification questions in
  grounded dialog.
\newblock In Houda Bouamor, Juan Pino, and Kalika Bali (eds.), \emph{Findings
  of the Association for Computational Linguistics: EMNLP 2023}, pp.\
  14988--14998, Singapore, December 2023. Association for Computational
  Linguistics.
\newblock \doi{10.18653/v1/2023.findings-emnlp.999}.
\newblock URL \url{https://aclanthology.org/2023.findings-emnlp.999/}.

\bibitem[{OpenAI}(2026)]{openai2026gpt56luna}
{OpenAI}.
\newblock {GPT-5.6 Luna} model.
\newblock Official API documentation, 2026.
\newblock URL \url{https://developers.openai.com/api/docs/models/gpt-5.6-luna}.
\newblock Accessed September 26, 2026.

\bibitem[Qiu et~al.(2025)Qiu, Liu, Weller, and Sch{\"o}lkopf]{qiu2025oft}
Zeju Qiu, Weiyang Liu, Adrian Weller, and Bernhard Sch{\"o}lkopf.
\newblock Orthogonal finetuning made scalable.
\newblock In \emph{Proceedings of the 2025 Conference on Empirical Methods in
  Natural Language Processing}, pp.\  31946--31963, Suzhou, China, November
  2025. Association for Computational Linguistics.
\newblock \doi{10.18653/v1/2025.emnlp-main.1627}.
\newblock URL \url{https://aclanthology.org/2025.emnlp-main.1627/}.

\bibitem[Sale et~al.(2023)Sale, Caprio, and H{\"u}llermeier]{sale2023volume}
Yusuf Sale, Michele Caprio, and Eyke H{\"u}llermeier.
\newblock Is the volume of a credal set a good measure for epistemic
  uncertainty?
\newblock In \emph{Uncertainty in Artificial Intelligence}, pp.\  1795--1804.
  PMLR, 2023.

\bibitem[Sale et~al.(2024)Sale, Bengs, Caprio, and
  H\"{u}llermeier]{sale2024second}
Yusuf Sale, Viktor Bengs, Michele Caprio, and Eyke H\"{u}llermeier.
\newblock Second-order uncertainty quantification: A distance-based approach.
\newblock In Ruslan Salakhutdinov, Zico Kolter, Katherine Heller, Adrian
  Weller, Nuria Oliver, Jonathan Scarlett, and Felix Berkenkamp (eds.),
  \emph{Proceedings of the 41st International Conference on Machine Learning},
  volume 235 of \emph{Proceedings of Machine Learning Research}, pp.\
  43060--43076. PMLR, 21--27 Jul 2024.
\newblock URL \url{https://proceedings.mlr.press/v235/sale24a.html}.

\bibitem[Sale et~al.(2026)Sale, Kotelevskii, Panov, and
  H{\"u}llermeier]{sale2026aleatoric}
Yusuf Sale, Nikita Kotelevskii, Maxim Panov, and Eyke H{\"u}llermeier.
\newblock {The Aleatoric-Epistemic Dichotomy of Uncertainty is Meaningful and
  Indispensable for Machine Learning}, May 2026.
\newblock URL \url{https://ssrn.com/abstract=6783338}.
\newblock SSRN preprint.

\bibitem[Schweighofer et~al.(2025)Schweighofer, Aichberger, Ielanskyi, and
  Hochreiter]{schweighofer2024information}
Kajetan Schweighofer, Lukas Aichberger, Mykyta Ielanskyi, and Sepp Hochreiter.
\newblock On information-theoretic measures of predictive uncertainty.
\newblock \emph{UAI}, 2025.

\bibitem[Tian et~al.(2023)Tian, Mitchell, Zhou, Sharma, Rafailov, Yao, Finn,
  and Manning]{tian-etal-2023-just}
Katherine Tian, Eric Mitchell, Allan Zhou, Archit Sharma, Rafael Rafailov,
  Huaxiu Yao, Chelsea Finn, and Christopher Manning.
\newblock Just ask for calibration: Strategies for eliciting calibrated
  confidence scores from language models fine-tuned with human feedback.
\newblock In \emph{Proceedings of the 2023 Conference on Empirical Methods in
  Natural Language Processing}, pp.\  5433--5442, Singapore, December 2023.
  Association for Computational Linguistics.
\newblock \doi{10.18653/v1/2023.emnlp-main.330}.
\newblock URL \url{https://aclanthology.org/2023.emnlp-main.330/}.

\bibitem[Van~Amersfoort et~al.(2020)Van~Amersfoort, Smith, Teh, and
  Gal]{van2020uncertainty}
Joost Van~Amersfoort, Lewis Smith, Yee~Whye Teh, and Yarin Gal.
\newblock Uncertainty estimation using a single deep deterministic neural
  network.
\newblock In \emph{International conference on machine learning}, pp.\
  9690--9700. PMLR, 2020.

\bibitem[Vashurin et~al.(2025)Vashurin, Goloburda, Ilina, Rubashevskii, Nakov,
  Shelmanov, and Panov]{vashurin2025cocoa}
Roman Vashurin, Maiya Goloburda, Albina Ilina, Aleksandr Rubashevskii, Preslav
  Nakov, Artem Shelmanov, and Maxim Panov.
\newblock {CoCoA}: A minimum {Bayes} risk framework bridging confidence and
  consistency for uncertainty quantification in {LLM}s.
\newblock In \emph{Advances in Neural Information Processing Systems},
  volume~38, pp.\  106236--106281. Curran Associates, Inc., 2025.
\newblock \doi{10.52202/085713-3546}.
\newblock URL
  \url{https://papers.nips.cc/paper_files/paper/2025/hash/98d2006bf15f00e2ff293dfba4acb8b2-Abstract-Conference.html}.

\bibitem[Walha et~al.(2026)Walha, Gruber, Decker, Yang, Javanmardi,
  H{\"u}llermeier, and
  Buettner]{walha2026finegraineduncertaintydecompositioninlargelanguagemodelsaspectralapproach}
Nassim Walha, Sebastian~G. Gruber, Thomas Decker, Yinchong Yang, Alireza
  Javanmardi, Eyke H{\"u}llermeier, and Florian Buettner.
\newblock Fine-grained uncertainty decomposition in large language models: A
  spectral approach.
\newblock \emph{Proceedings of the AAAI Conference on Artificial Intelligence},
  40\penalty0 (31):\penalty0 26090--26098, 2026.
\newblock \doi{10.1609/aaai.v40i31.39811}.
\newblock URL \url{https://ojs.aaai.org/index.php/AAAI/article/view/39811}.

\bibitem[Xu et~al.(2025)Xu, Guo, He, Hu, He, Bai, Chen, Wang, Fan, Dang, Zhang,
  Wang, Chu, and Lin]{xu2025qwen25omni}
Jin Xu, Zhifang Guo, Jinzheng He, Hangrui Hu, Ting He, Shuai Bai, Keqin Chen,
  Jialin Wang, Yang Fan, Kai Dang, Bin Zhang, Xiong Wang, Yunfei Chu, and
  Junyang Lin.
\newblock {Qwen2.5-Omni} technical report.
\newblock \emph{arXiv preprint arXiv:2503.20215}, 2025.
\newblock \doi{10.48550/arXiv.2503.20215}.
\newblock URL \url{https://arxiv.org/abs/2503.20215}.

\end{thebibliography}
